\documentclass[sigconf]{aamas}

\usepackage{balance} %

\usepackage[all]{foreign} %
\usepackage[inline]{enumitem}
\usepackage{tikz}
\usepackage{subcaption}
\usepackage{pgfplots}
\pgfplotsset{compat=1.18}
\usetikzlibrary{arrows.meta, positioning, fit, calc}
\usepackage{longtable}
\usepackage{multirow}
\usepackage{float}
\usepackage{algorithm}
\usepackage[noend]{algpseudocode}
\usepackage{placeins}

\doi{QMRY9661}

\newif\ifarxiv
\arxivtrue

\ifarxiv
\setcopyright{none}
\makeatletter
\renewcommand\footnotetextcopyrightpermission[1]{}%
\makeatother
\else
\makeatletter
\gdef\@copyrightpermission{
	\begin{minipage}{0.2\columnwidth}
		\href{https://creativecommons.org/licenses/by/4.0/}{\includegraphics[width=0.90\textwidth]{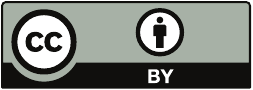}}
	\end{minipage}\hfill
	\begin{minipage}{0.8\columnwidth}
		\href{https://creativecommons.org/licenses/by/4.0/}{This work is licensed under a Creative Commons Attribution International 4.0 License.}
	\end{minipage}
	\vspace{5pt}
}
\makeatother

\setcopyright{ifaamas}
\acmConference[AAMAS '26]{Proc.\@ of the 25th International Conference
	on Autonomous Agents and Multiagent Systems (AAMAS 2026)}{May 25 -- 29, 2026}
{Paphos, Cyprus}{C.~Amato, L.~Dennis, V.~Mascardi, J.~Thangarajah (eds.)}
\copyrightyear{2026}
\acmYear{2026}
\acmDOI{}
\acmPrice{}
\acmISBN{}

\fi

\title[Scaling MEP through GNN]{Scaling Multi-Agent Epistemic Planning through GNN-Derived Heuristics}

\author{Giovanni Briglia}
\affiliation{
	\institution{Department of Sciences and Methods for Engineering, University of Modena and Reggio Emilia}
    \orcid{0000-0002-9572-3150}
	\city{Reggio Emilia}
	\country{Italy}}
\email{giovanni.briglia@unimore.it}
\additionalaffiliation{
	\institution{Department of Computer Science, University of Pisa}
	\city{Pisa}
	\country{Italy}}

\author{Francesco Fabiano}
\affiliation{
	\institution{Department of Computer Science, University of Oxford}
    \orcid{0000-0002-1161-0336}
	\city{Oxford}
	\country{United Kingdom}}
\email{francesco.fabiano@cs.ox.ac.uk}

\author{Stefano Mariani}
\affiliation{
	\institution{Department of Sciences and Methods for Engineering, University of Modena and Reggio Emilia}
    \orcid{0000-0001-8921-8150}
	\city{Reggio Emilia}
	\country{Italy}}
\email{stefano.mariani@unimore.it}

\begin{abstract}
	\emph{Multi-agent Epistemic Planning} (MEP) is an autonomous planning framework for reasoning about both the physical world and the beliefs of agents, with applications in domains where information flow and awareness among agents are critical.
	The richness of MEP requires states to be represented as \emph{Kripke structures}, \ie directed labeled graphs.
	This representation limits the applicability of existing heuristics, hindering the scalability of epistemic solvers, which must explore an exponential search space without guidance, resulting often in intractability.
	To address this, we exploit \emph{Graph Neural Networks} (GNNs) to learn patterns and relational structures within epistemic states, to guide the planning process.
	GNNs, which naturally capture the graph-like nature of Kripke models, allow us to derive meaningful estimates of state quality---\eg the distance from the nearest goal---by generalizing knowledge obtained from previously solved planning instances.
	We integrate these predictive heuristics into an epistemic planning pipeline and evaluate them against standard baselines, showing improvements in the scalability of multi-agent epistemic planning.
\end{abstract}

\keywords{Multi-Agent Epistemic Planning; Graph-Neural Networks; Learning in Planning; Heuristics}

\newcommand*{\EFP}{\mbox{EFP}\xspace}
\newcommand*{\HEFP}{\mbox{$\mathcal{H}$-\EFP}\xspace}

\newcommand*{\deep}{\texttt{deep}\xspace}

\newcommand*{\heurStyle}[1]{\texttt{#1}}
\newcommand*{\SUB}{\heurStyle{SUB}}
\newcommand*{\CPG}{\heurStyle{C\_PG}}
\newcommand*{\LPG}{\heurStyle{L\_PG}}

\newcommand*{\SPG}{\heurStyle{S\_PG}}

\newcommand*{\defemph}[1]{\ensuremath{\mathsf{#1}}}
\newcommand{\flstyle}[1]{\ensuremath{\mtt{#1}}}
\newcommand{\actstyle}[1]{\ensuremath{\mtt{#1}}}

\newcommand*{\agent}[1]{\ensuremath{\defemph{#1}}}%

\newcommand*{\opB}{\ensuremath{\mathbf{B}}}
\newcommand*{\opC}{\ensuremath{\mathbf{C}}}

\newcommand*{\cAlpha}[1]{\ensuremath{\opC_\alpha{#1}}}

\newcommand*{\bAg}[1]{\ensuremath{\opB_{\agent{#1}}}}
\newcommand*{\bB}[2]{\bAg{#1}\ensuremath{(#2)}}
\newcommand*{\cC}[2]{\ensuremath{\opC_{\agent{#1}}(#2)}}
\newcommand*{\del}{DEL}
\newcommand*{\brel}[1]{\ensuremath{\calB_{\agent{#1}}}}

\newcommand*{\mAL}{\mbox{\ensuremath{m\mathcal{A}^*}}}

\newcommand*{\mep}{MEP}

\newcommand*{\epg}{\emph{e}-PG\xspace}

\newcommand*{\lAG}{\ensuremath{\mathcal{L}_{\sAG}}}

\newcommand*{\lagC}{\ensuremath{\lAG^{\opC}}}

\newcommand*{\sAG}{\ensuremath{\mathcal{AG}}}
\newcommand*{\sAC}{\ensuremath{\mathcal{A}}}
\newcommand*{\sF}{\ensuremath{\mathcal{F}}}

\newcommand*{\Domain}{\ensuremath{D}}
\newcommand*{\Problem}{\ensuremath{P}}
\newcommand*{\ProblemArg}[1]{\Problem\ensuremath{(#1)}}

\newcommand*{\sFD}{\ProblemArg{\sF}}

\newcommand{\IniDomain}{\ensuremath{\calI}}
\newcommand{\GoalDomain}{\ensuremath{\calG}}

\newcommand*{\func}[3]{#1: #2 \mapsto #3}
\newcommand*{\shortimplies}{\ensuremath{\Rightarrow}}

\newcommand*{\bra}[1]{\ensuremath{\{#1\}}}
\newcommand*{\tuple}[1]{\ensuremath{\langle #1 \rangle}}

\newcommand*{\calB}{\ensuremath{\mathcal{B}}}

\newcommand*{\calG}{\ensuremath{\mathcal{G}}}

\newcommand*{\calI}{\ensuremath{\mathcal{I}}}

\newcommand{\his}{her\xspace}

\newcommand{\mtt}[1]{\ensuremath{\mathtt{#1}}}

\newcommand{\hyperstyle}[1]{\ensuremath{\mathsf{#1}}}
\newcommand{\resstyle}[1]{\texttt{#1}}
\newcommand{\bitmaskRes}{\resstyle{mask}\xspace}
\newcommand{\hashedRes}{\resstyle{hash}\xspace}
\newcommand{\mappedRes}{\resstyle{map}\xspace}

\newcommand{\BFSres}{\resstyle{BFS}\xspace}
\newcommand{\GNNres}{\resstyle{GNN}\xspace}

\newcommand{\statstyle}[1]{\hyperstyle{#1}}
\newcommand{\planLength}{\statstyle{Length}\xspace}
\newcommand{\solvingTime}{\statstyle{Time}\xspace}
\newcommand{\nodesExp}{\statstyle{Nodes}\xspace}
\newcommand{\IQM}{\statstyle{IQM}\xspace}
\newcommand{\IQR}{\statstyle{IQR}-\statstyle{std}\xspace}
\newcommand{\myAvg}{\statstyle{avg}\xspace}
\newcommand{\myStd}{\statstyle{std}\xspace}
\newcommand{\myTO}{\statstyle{TO}\xspace}
\newcommand{\unsolvedColumn}{\statstyle{-}\xspace}
\newcommand{\allInstances}{\statstyle{all}\xspace}
\newcommand{\onlyInCommon}{\statstyle{comm}\xspace}

\newcommand{\modelMinGR}{\statstyle{CC}-\statstyle{GR}\xspace}

\begin{document}

	\pagestyle{fancy}
	\fancyhead{}

	\maketitle

	\section{Introduction}

Planning scenarios involving multiple interacting entities, referred to as \emph{multi-agent}, have gained increasing importance due to their relevance in real-world applications, where groups of agents frequently need to interact.
However, effectively addressing multi-agent settings poses one of the most interesting challenges in modern AI research: adequately modeling multi-agent interaction while maintaining tractability~\citep{BRAFMAN201352}.
This is because such modeling requires accounting not only for the state of the world, but also for the dynamics of information exchange between agents.
Such reasoning, which deals with formalizing belief relationships among multiple agents, is referred to as \emph{epistemic reasoning}~\citep{fagin1994reasoning}.

Interest in \emph{Multi-agent Epistemic Planning} (MEP)---which integrates epistemic reasoning with automated planning---has surged~\citep{BBMvD17}, and several epistemic planners have been proposed~\citep{BolanderA11,muise2015planning,EngesserBMN17,KominisG17,huang2017general,icaps20,KR2021-12,Pham_Son_Pontelli_2023}.
To the best of our knowledge, only a few systems~\citep{icaps20,DBLP:conf/jelia/BuriganaFM23,Pham_Son_Pontelli_2023} are capable of reasoning over this setting without restrictions.
Nonetheless, these systems are severely limited by high computational costs, often making solving impractical.
This inefficiency stems mainly from two factors:
\begin{enumerate*}[label=\textit{(\arabic*)}]
    \item the intrinsic complexity of the underlying representations, which makes applying transitions and evaluating formulas within epistemic states (e-states) substantially harder than in classical planning; and
    \item the lack of effective heuristics, which results in a blind, combinatorial search as plan length increases.
\end{enumerate*}
While the aforementioned works in MEP largely address the first issue, few efforts tackle the latter.
A notable exception is the \HEFP\ planner~\citep{DBLP:conf/prima/FabianoPSP24}, an extension of~\citet{le2018efp}, which integrates heuristics guidance to improve scalability.
Our work builds upon this direction, sharing the core objective of designing effective heuristics extraction methods.
We argue this focus is essential, as \emph{informed search} is what enables scalability in planning systems---from classical heuristics planning~\citep{BONET20015,10.5555/1622559.1622565} to \emph{Monte Carlo Tree Search} (MCTS) in \emph{reinforcement learning} (RL) \citep{bouzy2006monte}.

The key difference in our approach lies in how heuristics are defined and computed.  
Unlike~\citet{le2018efp,DBLP:conf/prima/FabianoPSP24}, who construct heuristics using traditional planning constructs---such as the \emph{planning graph}---our method adopts a data-driven approach grounded in \emph{Machine Learning} (ML).  
Specifically, we leverage Graph Neural Networks (GNNs) to extract information from e-states in MEP---modeled as Kripke structures (Definition~\ref{def:kripke})---to estimate the ``quality'' of these states and derive heuristics functions.
The core idea is to use GNNs to approximate the \emph{perfect heuristics}, \ie to estimate the distance from any epistemic state to the nearest goal.
These learned heuristics are then used to guide an \emph{informed search} algorithm~\citep{BONET20015}, enabling efficient traversal of the search space and mitigating its exponential growth. %
We also introduce techniques for generating the training data required by the GNN-based regressor through a dedicated data generation process.
This entire pipeline is implemented in \deep\ (\texttt{d}ynamic \texttt{e}pistemic logic-bas\texttt{e}d \texttt{p}lanner),\footnote{
Code available at \url{https://github.com/FrancescoFabiano/deep}.} a novel iteration of the state-of-the-art epistemic planner \EFP~\citep{icaps20,DBLP:conf/prima/FabianoPSP24}.

The key contributions of this work are as follows:
\begin{enumerate}
    \item We define and compare three embeddings for Kripke structures to serve as input to a GNN-based regressor.
    \item We propose a fully automated pipeline for efficient data generation and training of the GNN-based regressor to approximate the perfect heuristics in the MEP setting.
    \item We integrate the GNN-regressor into the MEP solving process, where it is used to evaluate epistemic states by assigning heuristics scores that guide the search process.
    \item We provide a comprehensive evaluation of this integration by thoroughly testing several benchmarks.
\end{enumerate}
These contributions, supported by experimental results, represent a foundational step in integrating ML with MEP.

The remainder of this paper is structured as follows.
In Section~\ref{sec:background}, we provide background on MEP and GNNs.
Section~\ref{sec:core} presents our main theoretical contribution.
In particular, Section~\ref{sec:data-gen} presents the design of the embedding and the dataset generation while Section~\ref{sec:training-gnn} illustrates the training of the GNN-based regressor.
Section~\ref{sec:experiments} reports experimental results that evaluate the performance and scalability of our approach.
We discuss limitations and related work in Sections~\ref{sec:lim_and_future} and~\ref{sec:related}, and conclude in Section~\ref{sec:conc}.

	\section{Background}\label{sec:background}
\subsection{Dynamic Epistemic Logic}\label{subsec:del}
Dynamic Epistemic Logic (\del) formalizes reasoning about the state of the world and about the dynamic nature of information change, \ie about higher-order knowledge and/or beliefs.
For brevity, this discussion will present only the fundamental intuitions of \del.
Interested readers can explore further details in \citet{moss2015}.

Let us denote \sAG\ as a set of agents {such that $|\sAG|=n$ with $n \geq 1$}, and \sF\ as a set of propositional variables, referred to as \emph{fluents literals}, or simply \emph{fluents}.
Each \emph{world} is described by a subset of elements from \sF\ intuitively, those deemed True.
Furthermore, in epistemic logic, each agent $\agent{i} \in \sAG$ is associated to an epistemic modal operator $\bAg{i}$, signifying the belief\footnote{We use the terms knowledge and belief interchangeably, as their distinction is beyond this work’s scope. See \citet{fagin1994reasoning} for a full discussion.} of the agent.
Additionally, the epistemic \emph{group operator} \cAlpha\ is introduced.
Essentially, this operator represents the \emph{common knowledge} of a group of agents $\alpha$.

To be more precise, as in~\citet{baral2021action}, we have that a \emph{fluent formula} is a propositional formula built using fluents in \sF\ as propositional variables and the propositional operators $\wedge,\vee,\shortimplies,\neg$.
On the other hand, a \emph{belief formula} is either
\ifarxiv
\begin{itemize}
	\else
	\begin{enumerate*}[label=\textit{(\roman*)}]
		\fi
		\item a fluent formula;
		\item if $\varphi$ is a  belief formula and $\agent{i} \in \sAG$, then  $\bB{i}{\varphi}$ is a belief
		formula;
		\item if $\varphi_1, \varphi_2$ and $\varphi_3$ are belief formulae,
		then $\neg \varphi_3$ and $\varphi_1 \,\mathtt{op}\, \varphi_2$ are belief
		formulae, where $\mathtt{op} \in \bra{\wedge,\vee, \shortimplies}$; or
		\item if $\varphi$ is a belief formula and  $\emptyset \neq \alpha \subseteq \sAG$
		then $\cAlpha{\varphi}$
		is a belief formula.
		\ifarxiv
	\end{itemize}
	\else
\end{enumerate*}
\fi
\lagC\ denotes the language of the belief formulae over the set $\sAG$.

The classical way of providing semantics for epistemic logic is in terms of \emph{pointed Kripke structures}~\citep{Kripke1963-KRISCO}.

\begin{definition}[Pointed Kripke structure]\label{def:kripke}
	Let $|\sAG|=n$ with $n \geq 1$.
	A \emph{pointed Kripke structure} is a pair $(M=\tuple{S, \pi, \brel{1},\dots , \brel{n}},\defemph{s})$, such that:
	\begin{itemize}
		\item S is a set of worlds;
		\item $\func{\pi}{S}{2^{\sF}}$ is a function that associates an interpretation
		      of \sF\ to each element of S; %
		\item for $1 \leq \defemph{i} \leq \defemph{n}$, $\brel{i} \subseteq S \times S$  is a binary relation over S; and
  \item $\defemph{s} \in S$ points at the real world.
	\end{itemize}
\end{definition}
To elaborate, the component $S$ encompasses all the possible worlds configurations, while $\brel{i}$ specifically represents the beliefs held by each individual agent.

Intuitively, to verify whether a belief formula holds, we need to apply reachability within the Kripke model representing the e-state.
By exploring the set of reachable worlds obtained by applying epistemic operators, we determine which configurations of fluents an agent (or group of agents) considers possible.
Inconsistencies among these reachable worlds are used to model ignorance.
The formal semantics over pointed Kripke structures is provided in~\cite{icaps20,baral2021action} and omitted here as it is not integral to understanding the contribution of this paper.

\subsection{Multi-Agent Epistemic Planning}\label{subsec:mep}
We are now ready to introduce the fundamental concepts of MEP, while addressing interested readers to \citet{fagin1994reasoning,BolanderA11} for a more exhaustive introduction.

Let us begin by providing the notion of a \emph{multi-agent epistemic planning problem} in Definition~\ref{def:mMEP}.
Intuitively, an epistemic planning problem encompasses all the necessary information to frame a planning problem within a multi-agent scenario.

\begin{definition}[Multi-agent epistemic planning problem]\label{def:mMEP}
	We define a multi-agent epistemic problem as the tuple
	$\Problem=\tuple{\Domain = \tuple{\sF,\allowbreak\sAG,\allowbreak\sAC}\allowbreak,\IniDomain, \GoalDomain}$ where:
	\begin{itemize}%
		\item $\sF$ is the set of all the \emph{fluents} of \Problem;
		\item $\sAG$ is the set of  the \emph{agents} of \Problem;
		\item $\sAC$ represents the set of all the \emph{actions};

		\item $\IniDomain$ is the set of belief formulae that describes the \emph{initial conditions} of the planning process; and
		\item $\GoalDomain$ is the set of belief formulae that represents the \emph{goal} conditions.
	\end{itemize}
Note that the tuple $\Domain = \tuple{\sF,\sAG,\sAC}$ captures the domain description of which the problem $\Problem$ is an instance.

A solution, or a \emph{plan}, of a MEP problem is a sequence of actions in $\Domain$ that, when executed, transforms the initial e-state into one that satisfies the $\GoalDomain$.
\end{definition}

In this context, an epistemic state---represented by a pointed Kripke structure---encapsulates a problem's ``physical'' configuration along with the beliefs of the agents. %

To the best of our knowledge, the most widely accepted formalization of a comprehensive action language for multi-agent epistemic planning is \mAL\ \citep{baral2021action}.
Note that other languages capable of reasoning about DEL also exist~\citep{muise2015planning,KR2021-12}, but they limit their expressiveness in favor of efficiency and are therefore not considered here.

\mAL\ serves as a high-level action language facilitating reasoning about agents' beliefs within \lagC, where e-states are represented as Kripke structures.
It utilizes an English-like syntax, leverages \emph{event models} to define the transition functions, and uses reachability over Kripke models to characterize entailment.

In their work \citet{baral2021action} delineate three distinct types of actions within the context of multi-agent domains:
\begin{enumerate*}[label=\textit{(\roman*)}]
\item \emph{world-altering} actions (or \emph{ontic} actions): employed to modify specific properties, or fluents, within the world---denoted through the statement ``\actstyle{act\_name} \textbf{causes} $\mathtt{l}$\footnote{$\mathtt{l}$ can be either a fluent or its negation}'';
\item \emph{sensing} actions: used by an agent to refine \his beliefs about the world---denoted through the statement ``\actstyle{act\_name} \textbf{determines} \flstyle{f}''; and
\item \emph{announcement} actions: utilized by an agent to influence the beliefs held by other agents---denoted through the statement ``\actstyle{act\_name} \textbf{announces} \flstyle{f}''.
\end{enumerate*}
For brevity, we will not provide further details of \mAL\ here.
Interested readers can find a comprehensive description in \citet{baral2021action}.

\subsection{Graph Neural Networks (GNNs)}
Machine Learning consists of techniques that learn patterns from data and use them to make predictions and generalize, without being explicitly programmed.
GNNs \citep{bronstein2017geometric} extend this idea to graph-structured data: they update node representations by repeatedly exchanging information along edges.
This is possible through \textit{message-passing} mechanisms, which allow GNNs to detect structural and semantic regularities across nodes and subgraphs.
This makes GNNs particularly suitable for MEP, where epistemic states are represented as directed labeled graphs---\eg Kripke structures.

In our neural estimator, the GNN forms the first stage of the pipeline: it processes the graph representation of an epistemic state and produces a compact latent embedding that serves as the basis for the heuristic estimate, capturing the structural and logical characteristics of the state. Here, message passing is implemented via GINEConv~\citep{DBLP:conf/iclr/HuLGZLPL20}: given a node $v$ and its neighborhood $N(v)$, the update at layer $\ell$ is
\begin{equation*}
      h_v^{(\ell+1)}
  =
  \phi\!\Bigl(
    h_v^{(\ell)},
    \sum_{u \in N(v)}
      \psi\bigl(
        h_v^{(\ell)},\,
        h_u^{(\ell)},\,
        e_{uv}
      \bigr)
  \Bigr),
\end{equation*}
where $h_v^{(\ell)}$ is the embedding of node $v$ at layer $\ell$, $e_{uv}$ is the
embedding of the edge $(u,v)$, and $\psi$ and $\phi$ are multilayer perceptrons (MLPs). By comparison, the GCN \cite{DBLP:conf/iclr/KipfW17} layers update the node features while \emph{ignoring} the edge attributes, and the GAT \cite{DBLP:journals/corr/abs-1710-10903} layers introduce attention coefficients but still assume homogeneous edge types. The inclusion of $e_{uv}$ in $\psi$ makes GINE more expressive for relational structures such as Kripke structures, enabling the model to differentiate heterogeneous edge types (\eg distinct modal dependencies) that GCN and GAT cannot represent directly.

Reviews of architectures and applications are proposed in~\citet{wu2020comprehensive} and~\citet{zhou2020graph}, respectively.

	\section{Learning MEP Heuristics with GNN}
\label{sec:core}
As discussed, planning in MEP is an extremely resource-intensive task.
To mitigate this computational burden, we explore alternatives to \emph{Breadth-First Search} (\textbf{BFS}), such as \emph{Best-First Search}, which we abbreviate as \textbf{HFS} (for \emph{Heuristics-First Search}) to distinguish it from \textbf{BFS}.
\textbf{HFS} is a widely used strategy that prioritizes expanding states with the most favorable heuristics score.
Both \textbf{BFS} and \textbf{HFS} are standard approaches described in~\citet[Ch.~11]{modernApproach}.

\textbf{BFS} is an \emph{uninformed} search method, exploring the state space uniformly.
In contrast, \textbf{HFS} aims to guide the search more efficiently by prioritizing e-states that are likely closer to the goal.
The key challenge lies in defining an effective evaluation function---also known as a \emph{heuristics}---capable of accurately ranking epistemic states.
Heuristics are a central topic in the planning literature and have been extensively studied and proven effective~\citep{Keyder:2008:HPA:1567281.1567409,modernApproach}.
For this reason, a key contribution of this work is the formalization of heuristics tailored for \mep\ through the employment of GNNs.
{
}

\subsection{Training Data Generation}\label{sec:data-gen}
The first key challenge in employing ML techniques in \mep\ is determining what kind of information can be meaningfully extrapolated from data.
Several approaches have been proposed in automated planning, such as training models end-to-end using full problem descriptions and their associated goals~\citep{ijcai2023p839,huang2024planningdarkllmsymbolicplanning}.
However, these methods often suffer from low accuracy and require a large number of training instances to generalize effectively~\citep{kambhampati2024llmscantplanhelp}.

To address these limitations, we adopt a learning approach in which every state represents an independent training signal, rather than relying on complete trajectories.
This offers two main advantages.
First, although individual predictions may occasionally be inaccurate, their impact on the overall search is limited---as long as the trend of the heuristics is informative, the search remains effective.
Second, this approach significantly reduces the amount of data required.
Instead of needing complete problems as training instances, we can treat each e-state encountered during exploration as a distinct data point.
This allows the generation of tens of thousands of e-states, yielding a large set of training examples in a single run of the planner.

The main objective of this work is to use GNNs to extract information from the data to guide the search.
In particular, we aim at approximating the perfect heuristics that assign to each state its distance to the nearest goal.

\begin{figure*}[t]
    \centering
    \scalebox{0.99}{}\input{fig/architecture}
    \caption{
Illustration of the overall training and inference process.
On the left, we show dataset generation via DFS: 
teal nodes represent goals (score 0), black \emph{dotted} arrows show backtracking assigning distances, 
orange nodes (`x') indicate discarded branches, and gray nodes (`f') are states with no reachable goal.
Training is shown by the blue \emph{dashdotted} lines: $\langle$e-state, distance$\rangle$ pairs generated by the DFS are fed into the GNN to learn the properties of the e-states.
Following the magenta \emph{dashed} lines, we illustrate Inference where a single e-state---shown in its expanded view---is input to the GNN to retrieve its estimated distance to the goal.
The teal portion represents the goal encoding, while the magenta portion represents the actual e-state.
}
    \label{fig:dfs_generation}
    \Description{Illustration of the overall training and inference process.}
\end{figure*}

The need for reasoning over e-states is precisely why we adopt GNNs over other neural architectures.
As previously noted, e-states in \mep\ are represented as labeled, directed graphs, specifically, Kripke structures.
These grow unboundedly in size with the length of the plan.
In fact, although the set of possible world valuations is finite, \ie for a planning problem \Problem\ the number of valuations is exactly $2^{|\sFD|}$, the same valuation may appear multiple times within a Kripke model to capture complex belief relations.
This unbounded nature and the inherently relational semantics of epistemic states demand the full expressive power of graph-based models in order to capture the nuances of knowledge and belief.

\subsection{e-State Representation}\label{subsec:estate_repr}

The next challenge we address is designing a suitable data representation for integration within a GNN.
Each e-state $(M,s)$ is formalized as a pointed Kripke structure (Definition~\ref{def:kripke}), namely
\(
(M = \langle S, \pi, \brel{1}, \dots, \brel{n} \rangle, s).
\)

Encoding the relational structure \(\brel{i}\) is straightforward, as each agent \(\agent{i}\) can be mapped to a unique integer label.
The primary challenge lies in defining a consistent and informative numerical representation of the worlds \(S\) suitable for GNN input.

We explored three strategies, each offering a different trade-off between information preservation and efficiency.
In Section~\ref{subsec:ablation}, we compare these representations, illustrated below, and show that hash-based encoding achieves the best performance.

\paragraph{Symbolic ID Mapping.}
Each world \(s \in S\) is assigned a symbolic integer identifier through a mapping
\(
\phi : S \rightarrow \mathbb{N},
\quad \text{where} \quad 
\phi(s_i) = k \;\text{iff}\; s_i \text{ is the } k\text{-th distinct world encountered.}
\)
Intuitively, this assigns consecutive integers to newly observed worlds.
However, since this assignment depends on the order in which worlds are generated, \(\phi\) is not invariant across runs, leading to inconsistencies between datasets.
This representation has the lowest computational burden for data generation, training, and inference, but it incurs the greatest information loss, as similarities between worlds are not preserved.

\paragraph{Hash-based Encoding.}
A hash function
\(
h : S \rightarrow \mathbb{N}_{\ge 0}
\)
is applied to each world:
\(
h(s) = \mathrm{Hash}(\pi(s), r_s)
\);
where \(r_s\) denotes the repetition index distinguishing identical copies of the same world within an e-state.
This ensures consistent world identifiers across runs, but relative distances between world evaluations are lost, as, for example, evaluations differing only by a single fluent value may receive completely unrelated hash values.
We use the \texttt{hash\_range} function from the Boost libraries~\citep{10.5555/2049814} due to its high performance and existing use in \deep for e-state storage, which minimizes computational overhead.
Alternative hashing approaches may offer further improvements and are left for future work.

This encoding represents a middle ground in terms of the trade-off between information loss and the computational power required.

\paragraph{Bitmask Encoding.}
Each world \(s\) is represented as a purely binary vector:
\(
\mathbf{b}(s) = [\,\mathbf{r}_s \;|\; b_1, b_2, \dots, b_{l}\,],
\)
where \(\mathbf{r}_s\) is the binary representation of the repetition index \(r_s\),  
and \(b_j = 1\) if fluent \(\flstyle{f}_j \in \sF\) holds in \(\pi(s)\), and \(b_j = 0\) otherwise (or when \(j > |\sF|\)).
Note that, by construction, \(l\) is assumed to be greater than \(|\sF|\).
Thus, \(\mathbf{b}(s)\) encodes both the repetition number of the world within the e-state and its truth assignment over \(\sF\) as a single binary vector.
This embedding preserves most of the logical information about the e-estate, but it comes at the cost of a larger encoding, which requires more data for the model to converge appropriately and results in slower generation, training, and inference.

\subsubsection{Goal Encoding within the State Embedding}

While the encoded Kripke model---represented through the \emph{dot} language~\citep{Ellson2004}---proved effective, we encountered a second challenge.
Training solely on states led the regressor to learn absolute measures of distance.
This measure is independent of the underlying goal and therefore generalizes only to problems with identical goal conditions, which limits the applicability of the heuristics.

To mitigate this limitation, we extend the state embedding to include a compact encoding of the goal.
This enables the regressor to learn dependencies not only between the e-state and its distance to the goal, but also between structural features of the goal itself.
The design objective was to preserve a uniform input dimensionality while introducing minimal overhead.
Given that the conversion process is not particularly relevant, we omit its details here and refer the reader to 
\ifarxiv
 Appendix~\ref{app:goal-enc} 
\fi
for the description of the procedure.

In the resulting formulation, two types of nodes are encoded for each e-state embedding:
\ifarxiv
\begin{itemize}
\else
\begin{enumerate*}[label=\textit{(\roman*)}]
\fi
    \item \emph{state nodes}, corresponding to worlds in the epistemic state;
    \item \emph{goal nodes}, corresponding to symbols describing the goal conditions.
\ifarxiv
\end{itemize}
\else
\end{enumerate*}
\fi
This is exemplified in Figure~\ref{fig:dfs_generation}, where the left portion of the expanded hashed-based e-state (in teal) shows the goal encoding, and the right portion (in magenta) depicts the e-state itself.
The two are connected via a shared graph structure using nodes and edges that use constants and special identifiers throughout the process.

In the hashed- and in the mapping-based representations, the state component remains unchanged, while goal nodes are associated with unique integer identifiers (assigned to the first occurrences of each goal symbol).
In the bitmask representation, both node types share the same binary vector structure:
\(
\mathbf{b}(s) = [\, \mathbf{m}_s = [\, \mathbf{r}_s \;|\; b_1, b_2, \dots, b_{|l|} \,] \;|\; \mathbf{g}_s \,],
\)
where $\mathbf{m}_s$ reflects the bitmask embedding presented above and \(\mathbf{g}_s\) is the binary segment reserved for goal encoding.
During the experimental evaluation, we fixed the total size of $\mathbf{b}(s)$ to 64 bits, with $|\mathbf{r}_s| = 16$, $l = 32$, and $|\mathbf{g}_s| = 16$.

For \emph{state nodes}, the goal segment is not used (\(\mathbf{g}_s = \mathbf{0}\)), serving purely as padding to preserve dimensional uniformity across the GNN input space.
For \emph{goal nodes}, the segment \(\mathbf{g}_s\) encodes the binary representation of the integer identifier associated with the corresponding goal symbol while $\mathbf{m}_s = 0$.
Each goal symbol is uniquely represented, while agent-related components reuse the same integer indices as in the e-state encoding.

\subsection{Building the Dataset}
Having identified the type of data required for training, the next challenge lies in generating such data.
To address this, we equipped \deep with a \emph{dataset generation} mode that produces pairs of epistemic states and their distances to the nearest goal.
This allows training data to be collected from a small set of problems, which is then used to train GNN-based neural regressors.

The generation process works as follows: given a \mep\ problem, the planner performs a depth-limited \emph{Depth-First Search} (\textbf{DFS}) to explore the reachable state space up to a specified depth $d$ (left part of Figure~\ref{fig:dfs_generation}).
During this traversal, all reachable goal states are identified.
\deep then backtracks from each goal, assigning to each epistemic state the distance to the closest goal---yielding a dataset that approximates the perfect heuristics.
States from which no goal is reachable within depth $d$ are labeled with a special value.

Although conceptually simple, this process suffers from combinatorial explosion: with only 10 actions and depth 25---an overestimate of typical plan length in standard \mep\ benchmarks---the search space can reach $10^{25}$ nodes, making exhaustive exploration infeasible.
To mitigate this, we draw on ideas from \emph{local search}~\citep{Gerevini_2003}, sampling subregions to maximize coverage.
Our \textbf{DFS} incorporates probabilistic branch pruning (adaptive to depth and node count), randomized action ordering to avoid prefix bias, a node expansion cap, and duplicate e-state checks.
These mechanisms enable diverse yet tractable exploration, allowing informative datasets to be built within minutes per instance.

While data generation and training are handled offline in this work, the structure of our learning pipeline naturally supports an online setup.
By slightly extending the search process, the planner could incrementally collect training pairs and update the GNN once a sufficient number of samples is accumulated.
Thanks to \deep's multithreading support---introduced to emulate the portfolio behavior of \HEFP~\citep{DBLP:conf/prima/FabianoPSP24}---this online learning loop could run in parallel, allowing the planner to adapt and improve over time, similar to the cognitive architecture presented in~\citet{10.1145/3715709}.
We leave the exploration of this online learning paradigm to future work.

\subsection{Training Neural Distance Estimator}\label{sec:training-gnn}

Our objective is to train an estimator that, given an e-state, predicts the distance to a goal state.
To this end, three design choices were made:
\begin{enumerate}
    \item \textit{Discard unreachable nodes.} Distance to the goal is meaningful only for e-states from which the goal can be reached.
    Keeping unreachable e-states in the training set injects spurious labels that raise the estimator’s variance without lowering its bias.

    \item \textit{Limit the number of samples for any distance class.} Raw rollouts produce a strongly skewed distribution, with many more short distance e-states than long distance ones, which can bias the estimator toward the majority class.
    We therefore limit each distance bin (in percentage) to at most \hyperstyle{p_{M}} of the dataset.
    This balance step
    \begin{enumerate*}[label={\textit{(\alph*)}}]
    \item reduces variance arising from class imbalance while keeping bias low;
    \item forces the network to allocate capacity uniformly throughout the distance spectrum, lowering worst‑case error and improving robustness.
    \end{enumerate*}

    \item \textit{Linearly normalize the distance target.} To stabilize training, we linearly normalize the true distance \(d\in[0,\hyperstyle{D_{max}}]\) from $\hyperstyle{min\_val} \in [0,1)$ to $\hyperstyle{max\_val} \in (0, 1]$, with $\hyperstyle{max\_val} \ge \hyperstyle{min\_val}$.
    Let
    \(
        \alpha = \frac{\hyperstyle{max\_val} - \hyperstyle{min\_val}}{\hyperstyle{D_{max}}}, \text{ and } \beta = \hyperstyle{min\_val},
    \) then the normalized target is
    \(
        \tilde d = \alpha\,d + \beta \;\in[0,1].
    \)
    In inference, we recover the original scale through
    \(
        d = \frac{\tilde d - \beta}{\alpha} \in [0, \hyperstyle{D_{max}}]
    \)
    
    This normalization bounds the regression target, yielding predictable gradient magnitudes, avoiding activation saturation (\eg sigmoid or tanh), and aligning with common weight‐initialization schemes.
\end{enumerate}

Our neural regressor is implemented in \texttt{PyTorch} \cite{imambi2021pytorch}, and its graph encoder is built using \texttt{PyTorch} \texttt{Geometric} \cite{bielak2022pytorch}.
For each epistemic state (e-state), represented as a graph $G = (V, E)$ with nodes $V$ and edges $E$, we construct a corresponding data object that encodes the graph structure. This object serves as input to the GNN, which processes it to produce a latent embedding.
Specifically, this object includes the following:
\ifarxiv
\begin{itemize}
\else
\begin{enumerate*}[label=\textit{(\roman*)}]
\fi
    \item \textit{Node identifiers:} $V = \{v_i\}_{i=1}^{|V|}$, represented as bit-vectors, in which each node is expressed as a binary vector $b_i \in \{0,1\}^{d_b}$, where $d_b$ is the bit-length (here $d_b=64$).
    The resulting node feature matrix is $X \in \{0,1\}^{|V|\times d_b}$;
    \item \textit{Edge indices:} $E = \{(u_k, v_k)\}_{k=1}^{|E|}$, encoded as an index tensor $I \in \mathbb{N}^{2\times |E|}$, whose columns list each source–target pair;
    \item \textit{Edge attributes:} $A = \{a_k\}_{k=1}^{|E|},\quad a_k \in \mathbb{R}$, stacked into an attribute tensor $A \in \mathbb{R}^{|E|\times 1}$ and normalized to $[0,1]$.
\ifarxiv
\end{itemize}
\else
\end{enumerate*}
\fi
The estimator is trained by minimizing the Mean Squared Error ($\mathrm{MSE}$) between its predicted scalar $\hat d$ and the true normalized distance
\ifarxiv
\[\tilde d \in [0, 1]: \mathrm{MSE} \;=\;\frac{1}{B}\sum_{i=1}^B(\hat d_i - \tilde d_i)^2,\]
\fi
using the \texttt{AdamW} optimizer with the parameters listed in 
\ifarxiv
 Appendix~\ref{app:hyperparam}, 
\fi
 and where $B$ is the mini-batch size (number of samples per gradient update).

The forward pass unfolds in four stages:
\begin{enumerate}
    \item \textit{Bitwise embedding of node identifiers and edge attributes:} Each node bit-vector $b_i \in \{0,1\}^{d_b}$ is projected through a two-layer MLP,
    \ifarxiv
    \[
    x_i = f_\text{id}(b_i) = \mathrm{MLP}_{\text{id}}(b_i) \in \mathbb{R}^{d_v},
    \]
\fi

    producing dense embeddings of dimension $d_v=\hyperstyle{node\_emb\_dim}$.
    Similarly, each edge attribute $a_k$ is processed through
        \ifarxiv
    \[
    e_k = f_\text{edge}(a_k) = \mathrm{MLP}_{\text{edge}}(a_k) \in \mathbb{R}^{d_e},
    \]
\fi

    yielding $\hyperstyle{edge\_emb\_dim}$-dimensional embeddings.
    This bitwise encoding provides a compact and lossless representation of node identifiers while supporting continuous optimization and stable gradients.

    \item \textit{Relational message passing with GINEConv:} Two successive GINEConv \citep{DBLP:conf/iclr/HuLGZLPL20} layers, each followed by a ReLU activation, allow each node to aggregate information from its neighbors while explicitly incorporating edge indices and attributes.

    \item \textit{Graph-level summarization:} The node embeddings from the final GINE layer are aggregated into a fixed-size representation via \textit{mean pooling}:
        \ifarxiv
    \[
    h_G = \frac{1}{|V|}\sum_{v\in V} h_v.
    \]
\fi

    This operation is permutation-invariant and scale-independent, ensuring stable representations across graphs of different sizes.
    In contrast, sum pooling scales with $|V|$, potentially biasing larger graphs, while attention pooling introduces learnable weights but increases variance and computational cost.
    Mean pooling therefore offers the best generalization–variance trade-off for distance regression.

    \item \textit{Deep residual regression head with bounded output:} The pooled graph embedding $h_G$ is processed by a deep residual regressor:
    \(
    \hat d = \sigma\!\big(f_\text{res}(h_G)\big),
    \)
    where $f_\text{res}$ denotes a sequence of ResidualBlocks (Linear $\rightarrow$ BatchNorm $\rightarrow$ ReLU $\rightarrow$ Dropout $\rightarrow$ Linear $+$ skip).
    The sigmoid $\sigma$ bounds the output in $(0,1)$, and $\hat d$ is finally clamped to $[\hyperstyle{min\_val}, 1-\hyperstyle{min\_val}]$ to prevent numerical instabilities near the boundaries.
\end{enumerate}

Overall, the combination of \emph{bitwise node embeddings}, \emph{GINEConv} message passing, and \emph{mean pooling} yields a compact and expressive model that generalizes across variable-sized epistemic graphs while remaining numerically stable during training.
The full forward-pass is detailed in 
\ifarxiv
 Appendix~\ref{app:algorithm}.
\fi

	\section{Experiments and Evaluation}
\label{sec:experiments}
To conduct our experiments, we developed \deep, a modernized re-implementation of the epistemic planner \EFP~\citep{icaps20,DBLP:conf/prima/FabianoPSP24}.
\deep\ serves as the primary platform for our evaluation and features a modular design that supports the integration of diverse heuristics, including GNN–based ones.
Full implementation details and usage instructions are available at \url{https://github.com/FrancescoFabiano/deep}.

Here we present a comparative analysis of our primary contribution, \deep equipped with \textbf{HFS}$^{*}$---a depth-augmented variant of \textbf{HFS} detailed in Section~\ref{subsec:ablation}---alongside GNN-based heuristics (denoted as \GNNres), and \deep using Breadth-First Search (denoted as \BFSres).
All executions employed bisimulation-based state reduction and visited-state checks, as introduced by~\citet{icaps20}.
For completeness, we also provide comparisons with the different heuristics implemented in \HEFP~\citep{DBLP:conf/prima/FabianoPSP24}.
All of these heuristics remain available in \deep and can be combined with the GNN-based one in a portfolio approach.
The solver is already equipped to perform this integration easily as a runtime option.
We omit the results of such integration, as the focus of this paper is on the use of ML in epistemic planning, rather than on the planner itself.

While several other MEP solvers exist~\citep{muise2015planning,icaps20,KR2021-12}, we focus our evaluation on \BFSres to emphasize the impact of incorporating learned heuristics guidance into MEP solving.
All experiments were conducted with a timeout of 600 seconds on a 13th-generation Intel® Core™ i9-13900H CPU with 64 GB of system RAM and an NVIDIA RTX 4070 GPU with 8 GB of VRAM.

The evaluation encompasses several standard benchmarks in the MEP setting.
For brevity, the definitions of the parameters used for training and the descriptions of the domains, are provided in 
\ifarxiv
 Appendix~\ref{app:hyperparam} and~\ref{app:domains}, respectively.
\fi

\paragraph{Experimental Setup.}
To evaluate our contribution, we conducted experiments on standard benchmark domains.
We tested two configurations: in the first, each domain had its own model trained solely on data from that domain; in the second, we evaluated the knowledge transfer capabilities of models trained on data pooled from multiple domains and tested on both seen and unseen domains.

We report a subset of experiments to highlight key trends and the impact of our contribution, while full results are provided in 
\ifarxiv
 Appendix~\ref{app:experiments}.
\fi

\paragraph{Metrics.}

Our primary evaluation metric is the number of nodes expanded during the search (\nodesExp), which reflects the informativeness of the heuristics.
We focus on this measure because \GNNres currently does not leverage batch computation, making runtime comparisons less meaningful.
For completeness, we also report plan length (\planLength) and solving times in ms (\solvingTime).

For aggregate metrics, we report the Interquartile Mean (\IQM) and IQR standard deviation (\IQR) \citep{DBLP:conf/nips/AgarwalSCCB21}, focusing on problems solved by all approaches.

\subsection{E-State Representation and Search Strategy}\label{subsec:ablation}

In this section, we analyze how the e-state representation and the search strategy affect the overall performance of our approach.

Specifically, we compare the three e-state representations introduced in Section~\ref{subsec:estate_repr} and the two search strategies, \textbf{HFS} and its variant \textbf{HFS}$^{*}$. 
The latter, inspired by the classical \textbf{A}$^{*}$ algorithm\footnote{Since the GNN-based heuristics is statistical in nature, its admissibility cannot be guaranteed; hence we do not refer to it as \textbf{A}$^{*}$.}~\citep{FOEAD2021507,4082128}, augments the heuristics value $h(s)$ predicted by the GNN with the search depth $d(s)$ to form the total score:
\(
f(s) = h(s) + d(s).
\)

We evaluated all six possible combinations, namely the three representations, each paired with both \textbf{HFS} and \textbf{HFS}$^{*}$.
These experiments use GNN models trained on the same domain used during inference.
We report aggregate results over both the Training and Test sets.
We refer the reader to
\ifarxiv
 Appendix~\ref{app:experiments:ablation} 
\fi
for complete results.
We remind that the aggregates are computed only on instances solved by all the approaches.

Table~\ref{tab:ablation:datatype} reports results for the three e-state representations---\mappedRes, \hashedRes, and \bitmaskRes---corresponding respectively to the symbolic ID, hashing, and bitmask formulations.
As shown in Table~\ref{tab:ablation:datatype}, \hashedRes achieves the best overall performance.
This can be attributed to the intrinsic richness of the e-state representation, which enables effective information retrieval even when part of the original information is discarded.
Although the bitmask representation is only slightly outperformed by the hashed one in specific cases, we consider the trade-off offered by \hashedRes---lighter training requirements, faster data generation, and reduced inference time---well justifies choosing this as the best encoding.

\begin{table}[!ht]
  \centering
  \begin{tabular}{lcccc}
  	\toprule
      & Solved Inst.
      & \nodesExp
      & \solvingTime [ms]
      & \planLength \\
    \midrule
        \mappedRes    & 74/79 (93.67\%) & 87 $\pm$ 367 & 2358 $\pm$ 9251 & {6 $\pm$ 3} \\
    \hashedRes   & {75/79 (94.94\%)} & {45 $\pm$ 227} & {871 $\pm$ 3225} & {6 $\pm$ 3} \\
        \bitmaskRes    & {75/79 (94.94\%)} &  55 $\pm$ 224 & 1121 $\pm$ 4151 & {6 $\pm$ 3} \\
    \bottomrule
  \end{tabular}
\caption{Comparison of the three data representations.}
  \label{tab:ablation:datatype}
\end{table}

Next, in Table~\ref{tab:ablation:searchtype}, we illustrate the advantage of using \textbf{HFS}$^{*}$ over \textbf{HFS}.
The results clearly indicate that \textbf{HFS}$^{*}$ achieves better performance, solving approximately $32\%$ more instances.

\begin{table}[!ht]
  \centering
  \begin{tabular}{lcccc}
  	\toprule
      & Solved Inst.
      & \nodesExp
      & \solvingTime [ms]
      & \planLength \\
    \midrule
    \textbf{HFS}   & 49/79 (62.03\%) & 18 $\pm$ 27 & 1290 $\pm$ 5528 &  8 $\pm$ 8 \\
    \textbf{HFS}$^*$    & {75/79 (94.94\%)} & {17 $\pm$ 42} & {584 $\pm$ 2276} & {6 $\pm$ 4} \\
    \bottomrule
  \end{tabular}
    \caption{Comparison of \textbf{HFS} and \textbf{HFS}$^{*}$.}
  \label{tab:ablation:searchtype}
\end{table}

As mentioned, for the sake of space, we presented only aggregate results. 
We note, however, that the trends shown in Tables~\ref{tab:ablation:datatype} and~\ref{tab:ablation:searchtype} were consistently observed across all individual experiments. 
From this point onward, we therefore adopt \hashedRes combined with \textbf{HFS}$^{*}$ throughout the paper as our \GNNres configuration.

\subsection{Experimental Results}
Table~\ref{tab:batch1} summarizes the aggregate results across all domains, reporting the \IQM of \nodesExp, \solvingTime, and \planLength for each. It compares the GNN-based regressor (\GNNres) with uninformed search (\BFSres), where the GNN models are trained on the same domain used at inference time. We report only the results of the Test set (\ie, problems not seen during training).

\begin{table*}[!ht]
  \centering
  \begin{tabular}{lcccc|cccc}
  	\toprule
    & \multicolumn{4}{c|}{\GNNres} &
    \multicolumn{4}{c}{\BFSres} \\
    \midrule
      & Solved Inst.
      & \nodesExp
      & \solvingTime [ms]
      & \planLength
      & Solved Inst.
      & \nodesExp
      & \solvingTime [ms]
      & \planLength\\
    \midrule
    \textbf{AL} & {6/7 (85.71\%)} &  10 $\pm$ 0 & 274 $\pm$ 3142 & {5 $\pm$ 0}  & {6/7 (85.71\%)} &  14 $\pm$ 0 & 82 $\pm$ 3851 & {5 $\pm$ 0} \\
    \textbf{CC} & {18/18 (100.00\%)} &  65 $\pm$ 250 & 996 $\pm$ 2006 & {7 $\pm$ 4}  & {18/18 (100.00\%)} &  610 $\pm$ 1607 & 1610 $\pm$ 6323 & {5 $\pm$ 3} \\
    \textbf{CB} & {3/3 (100.00\%)} &  75 $\pm$ 860 & 279 $\pm$ 3509 & {5 $\pm$ 2}  & {3/3 (100.00\%)} &  102 $\pm$ 1260 & 118 $\pm$ 1603 & {5 $\pm$ 2} \\
    \textbf{GR} & {8/12 (66.67\%)} &  157 $\pm$ 380 & 6512 $\pm$ 14016 & {6 $\pm$ 2}  & {10/12 (83.33\%)} &  448 $\pm$ 2068 & 5094 $\pm$ 12086 & {4 $\pm$ 2} \\
    \textbf{SC} & {19/20 (95.00\%)} &  49 $\pm$ 357 & 157 $\pm$ 511 & {10 $\pm$ 6}  & {19/20 (95.00\%)} &  114 $\pm$ 372 & 131 $\pm$ 572 & {8 $\pm$ 4} \\
    \textbf{SR} & {5/6 (83.33\%)} &  6188 $\pm$ 17295 & 38111 $\pm$ 106636 & {8 $\pm$ 10}  & {5/6 (83.33\%)} &  7918 $\pm$ 22608 & 42146 $\pm$ 120182 & {8 $\pm$ 10} \\
\midrule
    \textbf{All} & 59/66 (89.39\%) &  64 $\pm$ 296 & 1001 $\pm$ 4635 & {7 $\pm$ 4}  & {61/66 (92.42\%)} &  242 $\pm$ 1080 & 922 $\pm$ 4870 & {6 $\pm$ 4} \\
    \bottomrule
  \end{tabular}
\caption{Comparison between \GNNres and \BFSres across standard benchmarks. \textbf{All} represents results over the entire Test set.}
  \label{tab:batch1}
\end{table*}

In Table~\ref{tab:aggregate}, as a study on the knowledge transfer analysis, we compare \GNNres with  \modelMinGR, a model trained using the training instances of two domains, namely \textbf{CC} and \textbf{GR}.

Table~\ref{tab:batch4-hefp} compares our approach with existing heuristics in \HEFP. 
For this evaluation, we use \modelMinGR, as it represents the most general-purpose configuration. 
A detailed comparison with individual heuristics (\CPG, \LPG, \SPG, and \SUB) and their description is provided in
\ifarxiv
 Appendix~\ref{app:experiments}.
\fi
We note that the only metric with significant interpretability is the number of instances solved, since the \HEFP\ heuristics solve a highly diverse set of instances, making aggregated measures less informative.

Tables~\ref{tab:aggregate} and~\ref{tab:batch4-hefp} summarize performance on the full \textit{Test} set across all benchmark domains.

\begin{table}[!ht]
  \centering
  \begin{tabular}{lcccc}
  	\toprule
      & Solved Inst.
      & \nodesExp
      & \solvingTime [ms]
      & \planLength \\
    \midrule
    {\BFSres}   & 55/59 (93.22\%) & 389 $\pm$ 1410 & 1384 $\pm$ 6943 &  6 $\pm$ 4 \\
    {\modelMinGR}    & 55/59 (93.22\%) & 288 $\pm$ 1244 & 4116 $\pm$ 13644 & 6 $\pm$ 4 \\
    \bottomrule
  \end{tabular}
    \caption{Comparison between \GNNres equipped with \modelMinGR and \BFSres across standard benchmarks.}
  \label{tab:aggregate}
\end{table}

\begin{table}[!ht]
\centering
\begin{tabular}[!ht]{lrr}
\toprule
Approach & \# Solved & \% Solved \\
\midrule
\GNNres & 64/75  & $85.33\%$ \\
\CPG & 37/75  & $49.33\%$ \\
\LPG & 54/75 & $72.00\%$ \\
\SPG & 62/75 & $82.67\%$ \\
\SUB & 58/75 & $77.33\%$ \\
\bottomrule
\end{tabular}
\caption{\modelMinGR against \HEFP's individual heuristics.}
\label{tab:batch4-hefp}
\end{table}

\subsection{Discussion}

\GNNres demonstrates informative and robust performance across a range of planning domains, as evidenced by the aggregate metrics reported in Table~\ref{tab:batch1}.
With the exception of the \textbf{GR} domain, \GNNres consistently reduces the number of explored nodes compared to uninformed search, highlighting its efficiency gains.

The \textbf{GR} domain presents a particular challenge due to its sparse solution density: even small heuristic inaccuracies can result in poor search guidance, and instances with superficially similar states or goals may require substantially different plans.
This variability can mislead \GNNres and explains its reduced effectiveness in this domain.
Importantly, this observation motivates the portfolio-based heuristic selection strategy adopted in \deep, where complementary heuristics can compensate when individual ones underperform.
A tighter integration of GNN-based heuristics into this portfolio framework is a natural direction for future work.

Further evidence of scalability is provided in Table~\ref{tab:aggregate}, where \GNNres outperforms the baseline in terms of explored nodes, demonstrating its ability to generalize across domains, including those unseen during training. We focus on expanded nodes as this metric directly reflects heuristic informativeness rather than implementation dependent overhead.
At this stage, GNN inference remains a proof of concept and lacks several engineering optimizations, \eg batch inference, making runtime performance non-competitive (more information can be found in \ifarxiv
 Appendix~\ref{app:batch-estimate}).
\fi
In contrast, heuristics informativeness is a stable property, expected to persist once such optimizations are introduced.

Results for \planLength are comparable across methods, indicating that \GNNres preserves near-optimal solution quality.
Solving times are likewise similar, although \GNNres incurs a modest overhead due to the absence of batch computation.
A detailed analysis of how batch inference would affect \GNNres---bringing its per-node overhead closer to that of \BFSres---is presented in
\ifarxiv
 Appendix~\ref{app:batch-estimate}.
\fi

Table~\ref{tab:batch4-hefp} further compares \GNNres against individual heuristics within \HEFP.
\GNNres performs on par with the best individual heuristics, demonstrating its viability as an alternative approach.
At the same time, we view GNN-based heuristics as a complementary tool to enhance solver coverage, \eg through integration into portfolio-based approaches, rather than as a replacement for existing heuristics.

Overall, \GNNres achieves consistent reductions in explored nodes and represents a scalable, learning-based step toward effective heuristics for multi-agent epistemic planning, a setting in which heuristic guidance is currently limited.

	\section{Limitations and Future Directions}
\label{sec:lim_and_future}
While our approach achieves promising experimental results, several limitations remain.
First, we acknowledge that our current implementation does not yet achieve competitive runtime performance compared to existing heuristics methods.
This limitation is largely due to engineering considerations.
A proper integration of CUDA-based computation and batch processing, combined with a search strategy capable of exploiting these features while minimizing memory exchange, would, in fact, significantly improve inference speed as detailed in 
\ifarxiv
 Appendix~\ref{app:batch-estimate}.
\fi
Although we recognize this shortcoming, we emphasize that this work is foundational.
Addressing the engineering challenges required for an optimized implementation would constitute a substantial effort in its own right---worthy of dedicated study---and represents an important avenue for future research.
Nonetheless, our primary goal here is to provide a proof of concept highlighting the potential of heuristics learning in MEP.
This is also why we focus on the number of expanded nodes as our primary evaluation metric.

Certain domains, such as \textbf{AL} and \textbf{GR}, also present unique challenges.
In \textbf{AL}, problem instances differ only in the nesting depth of belief formulas, which results in weak learning signals.
Results for this domain are therefore not very informative as \GNNres and \BFSres perform almost identically.
In \textbf{GR}, the sparsity of valid plans limits the effectiveness of data-driven learning, exposing a current limitation of our data generation pipeline as discussed above.

As mentioned, our current \GNNres implementation lacks {batch inference} during planning, which contributes to slower runtime.
While enabling batch computation is primarily an engineering task, it also raises design questions regarding when and how to accumulate batches of states for scoring.
For instance, the planner could rely on \textbf{BFS} or alternate with other heuristics until a sufficient number of candidate states are available for batched GNN evaluation.
We leave the systematic study of these strategies for future work.

Finally, an important next step is to integrate our GNN-based heuristic estimates into more advanced search frameworks, such as Monte Carlo Tree Search~\cite{sutton1998reinforcement}.
We believe this approach has strong potential to improve the scalability and adaptability of MEP solvers.

\section{Related Works}
\label{sec:related}
\paragraph{Machine Learning in Planning.}
{}

Traditionally, planning heuristics are either hand-crafted or derived from structural features of the search space~\citep[Ch.~11]{modernApproach}.
ML-based heuristics offer a scalable alternative, learning meaningful patterns from data~\citep{jimenez2012review,chen2024return}.
This is exemplified by systems like AlphaGo~\citep{silver2016mastering}, where learned guidance enables scalable MCTS.
Our work builds on these ideas but targets a more structured setting, where planning states are represented as Kripke structures.
This introduces challenges, which we address using GNNs to extract semantic features from epistemic states.
GNNs have also proven effective in classical planning, where they model relational graphs~\citep{silver2021planning}, learn numeric heuristics~\citep{borelli2025learning}, or guide adaptive search~\citep{du2025fast}.

Recent efforts have also explored using Large Language Models (LLMs) in planning.
While LLMs are ineffective as standalone planners~\citep{kambhampati2024llmscantplanhelp,DBLP:conf/icaps/PallaganiM0FLMS24}, they can aid heuristics generation~\citep{kambhampati2024llmscantplanhelp,correa2025classical} or domain formalization~\citep{tantakoun-etal-2025-llms}.
However, due to the structured nature of MEP, we believe GNNs are a more suitable choice.
We leave the integration of LLMs in this context to future work.

\paragraph{Multi-Agent Epistemic Planning.}
Most work on MEP has focused on foundational problems such as the investigation of DEL fragments~\citep{DBLP:conf/atal/BuriganaFMT23}, the definition of action languages~\citep{muise2015planning,baral2021action,DBLP:conf/aiia/BuriganaF22a}, and the development of underlying representations~\citep{icaps20,DBLP:conf/jelia/BuriganaFM23}.
While these are fundamental contributions, this paper pursues a different goal: enabling efficient exploration via informed search.

To the best of our knowledge, only one line of work addresses this challenge, namely~\citet{le2018efp,DBLP:conf/prima/FabianoPSP24}, which derive heuristics using the planning graph structure.
Our approach differs by employing data-driven methods.
As mentioned, a future direction is to investigate the interplay between these two types of heuristics either in parallel or in conjunction.

Other recent efforts integrate ML and RL with epistemic planning but simplify epistemic state representations.
\citet{engesser2025simple} decompose e-states into bounded feature vectors, bridging epistemic logic and reinforcement learning.
Similarly,~\citet{nunn2024logic} use generative models to reason over individual formulae rather than full Kripke structures, enabling different but interesting capabilities.

\section{Conclusions}
\label{sec:conc}
This work introduces a novel, learning-based approach to heuristics generation for multi-agent epistemic planning, leveraging Graph Neural Networks to guide informed search.
By embedding Kripke structures and training a GNN to approximate the perfect heuristics, we enable scalable MEP planning through learning.

We investigated three distinct embeddings of Kripke structures and evaluated their effect on the heuristics accuracy.
Through a comprehensive benchmarking, we identified the embedding configurations that most effectively balance expressiveness and scalability.

Our implementation, \deep, demonstrates solid performance across standard benchmarks, reducing the number of explored nodes compared to uninformed search.
The method also generalizes well to unseen domains and is competitive against existing heuristics.

These results highlight the potential of heuristics learning in MEP, where heuristics are scarce.

For these reasons, this work represents a foundational step toward exploiting machine learning in the context of multi-agent epistemic planning.

\begin{acks}
This research was partially funded by the EPSRC grant EP/Y028872/1, \emph{Mathematical Foundations of Intelligence: An ``Erlangen Programme'' for AI}.
\end{acks}

	\bibliographystyle{ACM-Reference-Format} 
	\bibliography{aamas2026}
	
	\ifarxiv
    	\clearpage
    	{\onecolumn

\appendix
\begin{center}
    {\LARGE \textbf{Technical Appendix}}
\end{center}

\vspace{4mm}

\section{Computational Resource Used}\label{app:comp-res}
\FloatBarrier
All experiments were performed on a $13^{th}$ Intel(R) Core(TM) i9-13900H with 20 CPU cores and 64 GB of system RAM, alongside an NVIDIA RTX 4070 GPU with 8 GB of dedicated VRAM.
Note that the experiments do not need multiple repetitons as the inference and the planning process does not depend on randomness.

Overall, for the final results, sample generation, model training and inference across all experimental batches required approximately 180 hours of high‑performance computing (HPC) time.
Since the project began, however, simulation trials have consumed nearly 400 hours.

Illustrative timing breakdown for a single model:
\begin{itemize}
    \item \textit{Sample generation:} Approximately 15 minutes per experimental setup.
    \item \textit{Data‑loader construction:} 15 minutes for fewer than 10k samples; up to 2 hours for more than 100k samples.
    \item \textit{Model training:} 10 minutes for very small data‑loaders; up to 3 hours for very large ones.
\end{itemize}

\section{Parameters and Hyperparameters}\label{app:hyperparam}
\FloatBarrier
In this section, we summarize all model and optimizer settings used in our experiments.
Table~\ref{tab:estimator_params} lists the architectural parameters of the \hyperstyle{GNN}-\hyperstyle{based}-\hyperstyle{Regressor}; while Table~\ref{tab:adamw_params} gives the key AdamW optimizer hyperparameters.

\label{sec:parameters}
\begin{table*}[!ht]
    \centering
    \begin{tabular}{c|c}
        \textbf{Parameter Description} & \textbf{Value} \\
        \hline
            Size of hidden feature dimension in GINEConvs and regressor input. & 128 \\
            Dimension of initial node embeddings (ID MLP output). & 64 \\
            Dimension of initial edge embeddings (edge MLP output). & 32 \\
            Hidden size in the residual regressor head. & 128 \\
            Number of ResidualBlock layers in the regressor. & 3 \\
            Dropout probability inside each ResidualBlock. & 0.2 \\
            Lower clamp bound on final sigmoid output. & 1e-3 \\
            Upper clamp bound on final sigmoid output. & 1 - 1e-3 \\
    \end{tabular}
    \caption{Values and descriptions of the \hyperstyle{GNN}-\hyperstyle{based}-\hyperstyle{Regressor} parameters.}
    \label{tab:estimator_params}
\end{table*}

\begin{table*}[!ht]
    \centering
    \begin{tabular}{c|c}
        \textbf{Hyperparameter Description} & \textbf{Value} \\
        \hline
        Learning rate for parameter updates. & $10^{-3}$ \\
        Exponential decay rates for the first and second moment estimates. & (0.9, 0.999) \\
        Term added to the denominator to improve numerical stability. & $10^{-8}$ \\
        Coefficient for decoupled weight decay (L2 regularization). & $10^{-2}$ \\
        Flag to enable the AMSGrad variant of AdamW. & False \\
    \end{tabular}
    \caption{Key hyperparameters of the AdamW optimizer.}
    \label{tab:adamw_params}
\end{table*}

\subsection{Case studies}

Here, an overview of the experimental configuration parameters is provided.

\begin{table}[!ht]
    \centering
    \small
     \begin{tabular}{c|c|c}
        \textbf{Parameter} & \textbf{Description} & \textbf{Value} \\
        \hline
        $\hyperstyle{D_{\max}}$ & Maximum goal distance considered & 50 \\
        \hline
        $\hyperstyle{d_{b}}$ & Length of the input bitmask vector & 64 \\
        \hline
        $\hyperstyle{min\_val}$ & Lower bound for linear normalization & $10^{-3}$ \\
        \hline
        $\hyperstyle{max\_val}$ & Upper bound for linear normalization & $1 - 10^{-3}$ \\
        \hline
        $\hyperstyle{p_{M}}$ & Maximum class imbalance threshold in the target distribution & 0.5 \\
    \end{tabular}
    \caption{Experiment configuration parameters: overview of each parameter and its respective value.}
    \label{tab:experimental_conf}
\end{table}

\clearpage
\section{Goal Encoding into e-State Embedding}\label{app:goal-enc}
\FloatBarrier
We first note that generating a Kripke structure corresponding to the goal state is not a viable solution.
As discussed by~\citet{10.1007/978-3-319-11558-0_17}, there exist infinitely many variations of epistemic states that satisfy a given arbitrary belief formula---\ie a goal description---making it impossible to generate all of them.
Generating only one or a few such states would bias the training process toward those specific representations, favoring e-states that structurally resemble the generated ones.
This is problematic, as the structure of valid e-states can vary significantly.

Moreover, as shown in~\citet{bolander2015complexity}, the computational complexity of generating such structures is exponential in complexity with respect to the size of the formula.

To address this, we developed a custom syntax tree that parses the problem's goal into an ad-hoc graph-based representation.
This representation highlights key components such as belief operators and logical connectives, while maintaining consistent naming conventions with the state representation.
Specifically, the only shared identifiers between the goal and the e-state are those that denote the same underlying entity---in our case, the agent IDs.

The procedure is presented in Algorithm~\ref{alg:goal-subtree}
\begin{algorithm}[H]
\caption{Recursive Goal Formula Graph Construction}
\label{alg:goal-subtree}
\begin{algorithmic}[1]
\Statex \textbf{Input:} Belief formula $\varphi$, goal ID $\mathtt{g}$, node counter $\mathtt{next\_id}$, parent node $\mathtt{p}$
\Statex \textbf{Output:} The graph is printed recursively on the output file

\Function{GenerateGoalEmbedding}{$\varphi,\, \mathtt{g},\, \mathtt{next\_id},\, \mathtt{p}$}
    \State $\mathtt{id} \gets \texttt{increment}(\mathtt{next\_id})$
    \State $\mathtt{n} \gets \texttt{string}(\mathtt{id})$
    
    \If{$\varphi$ is a fluent formula}
        \ForAll{subformula $\mathtt{S}$ in $\varphi$} \Comment{Handles logic OR}
            \If{$\mathtt{S}$ has multiple fluents} \Comment{Handles logic AND}
                \State $\mathtt{n} \gets \texttt{new\_node\_id}$
                \State connect $\mathtt{p} \rightarrow \mathtt{n}$ with label $\mathtt{g}$
                \State $\mathtt{p} \gets \mathtt{n}$
            \EndIf
            \ForAll{fluent $\mathtt{f}$ in $\mathtt{S}$}
                \State connect $\mathtt{p} \rightarrow \texttt{get\_f\_id}(\mathtt{f})$ with label $\mathtt{g}$ \Comment{Retrieves unique fluent ID}
            \EndFor
        \EndFor
    \ElsIf{$\varphi = \bB{a}{\psi}$}
        \State connect $\mathtt{p} \rightarrow \mathtt{n}$ with label $\mathtt{g}$
        \State connect $\mathtt{n} \leftrightarrow \texttt{get\_a\_id}(\mathtt{a})$ with label $\mathtt{g}$ \Comment{Retrieves unique agent ID}
        \State \Call{GenerateGoalEmbedding}{$\psi,\, \mathtt{g},\, \mathtt{next\_id},\, \mathtt{n}$}
    \ElsIf{$\varphi = \cC{G}{\psi}$}
        \State connect $\mathtt{p} \rightarrow \mathtt{n}$ with label $\mathtt{g}$
        \ForAll{agent $\mathtt{a}$ in group $\mathtt{G}$}
            \State connect $\mathtt{n} \leftrightarrow \texttt{get\_a\_id}(\mathtt{a})$ with label $\mathtt{g}$
        \EndFor
        \State \Call{GenerateGoalEmbedding}{$\psi,\, \mathtt{g},\, \mathtt{next\_id},\, \mathtt{n}$}
    \ElsIf{$\varphi$ is a propositional formula}
        \State connect $\mathtt{p} \rightarrow \mathtt{n}$ with label $\mathtt{g}$
        \State \Call{GenerateGoalEmbedding}{$\psi_1,\, \mathtt{g},\, \mathtt{next\_id},\, \mathtt{n}$}
        \If{$\psi_2$ exists}
            \State \Call{GenerateGoalEmbedding}{$\psi_2,\, \mathtt{g},\, \mathtt{next\_id},\, \mathtt{n}$}
        \EndIf
    \EndIf
\EndFunction
\end{algorithmic}
\end{algorithm}

\clearpage
\section{Domains}\label{app:domains}
\FloatBarrier
Here we present the complete description of the benchmarks used to evaluate our contributions.
These have been collected from the literature~\citep{kominis2015beliefs,huang2017general,cooper:hal-02147986,icaps20}.
\begin{itemize}
    \item \emph{Assembly Line} (\textbf{AL}). This domain involves two agents, each responsible for processing a separate part of a product.
    Each processing step may fail, and agents can inform one another of their task’s outcome.
    Based on this shared knowledge, the agents decide whether to \emph{assemble} the product or \emph{restart}.
    The goal is fixed—assembling the product—but the complexity varies depending on the \emph{depth} of the belief formulas used in the executability conditions.

    \item \emph{Collaboration and Communication} (\textbf{CC}). In this domain, $n \geq 2$ agents move along a corridor with $k \geq 2$ rooms, in which $m \geq 1$ boxes can be located.
    Whenever an agent enters a room, she can observe whether a specific box is present.
    Additionally, agents can communicate information about the boxes' positions to other \emph{attentive} agents.
    The goals involve both agents' physical positions and their beliefs about the boxes.

    \item \emph{Coin in the Box} (\textbf{CB}). Here, $n \geq 3$ agents are in a room with a locked box containing a coin, which lies either heads or tails up.
    Initially, no agent knows the coin's orientation.
    One agent holds the key to open the box.
    Typical goals involve some agents learning the coin’s status, while others may need to know that someone else knows it—or remain ignorant of this fact entirely.

    \item \emph{Grapevine} (\textbf{GR}). In this setting, $n \geq 2$ agents are distributed across $k \geq 2$ rooms.
    Agents can freely move between rooms and share "secrets" with any other agents present in the same room.
    This domain supports diverse goal types, ranging from secret sharing to creating misconceptions about others' beliefs.

    \item \emph{Selective Communication} (\textbf{SC}). This domain features $n \geq 2$ agents, each starting in one of the $k \geq 2$ rooms arranged along a corridor.
    An agent may broadcast information, which is heard by all agents in the same or adjacent rooms.
    Each agent can move between neighboring rooms.
    The goals often require certain agents to know specific facts while ensuring that others remain unaware of them.

    \item \emph{Selective Communication Enriched} (\textbf{SCR}). This domain is a replica of \textbf{SC} but enriched with extra, useless actions, to decrease the density of the solutions in the search space.

   \item \emph{Epistemic Gossip} (\textbf{EG} -- Only used in experiments presented in the Appendix). This domain extends the classic gossip problem, where information (``secrets") must be disseminated among $n \geq 2$ agents using the minimum number of calls.
    Unlike the traditional formulation, which requires some agents to know some secrets (epistemic depth 1), we consider arbitrary epistemic depths.
    For example, goals may require that all agents know that all agents know all secrets (depth 2), and so on.
    
\end{itemize}

\newpage
\section{Forward Pass Neural Distance Regressor}\label{app:algorithm}
\FloatBarrier

\begin{algorithm}[H]
\caption{Batch-wise Distance Estimation (bit-vector IDs, no goal, no depth)}
\label{algo:forward}
\begin{algorithmic}[1]
\Statex \textbf{Input:} $\mathtt{BatchDict}$ with entry:
\begin{itemize}
  \item \texttt{state\_graph} $= (\mathtt{B},\, \mathtt{E_i},\, \mathtt{E_a},\, \mathtt{batch})$
  \begin{itemize}
    \item $\mathtt{B} \in \{0,1\}^{N \times d_b}$: node bit-vectors ($d_b$ bits per node, typically $d_b = 48$);
    \item $\mathtt{E_i} \in \mathbb{N}^{2 \times |E|}$: edge indices in \textbf{COO format} (Coordinate format), where each column $(u, v)$ identifies a directed edge $u \!\rightarrow\! v$;
    \item $\mathtt{E_a} \in \mathbb{R}^{|E| \times 1}$: edge attributes (scalar per edge);
    \item $\mathtt{batch} \in \mathbb{N}^{N}$: batch assignment of each node to its graph.
  \end{itemize}
\end{itemize}
\Statex \textbf{Output:} $\mathbf{d} \in \mathbb{R}^b$, estimated distances per batch ($b$ = number of graphs)

\Function{Regressor}{$\mathtt{r}$}
  \State $\mathtt{h^{(0)}} \gets \mathrm{ReLU}\!\big(W^{(1)} \mathtt{r} + b^{(1)}\big)$
  \For{$j = 1$ to $BL$} \Comment{$BL$ = number of residual blocks}
    \State $\mathtt{u} \gets W^{(j,1)} \mathtt{h^{(j-1)}} + b^{(j,1)}$
    \State $\mathtt{u} \gets \mathrm{BN}_{j,1}(\mathtt{u})$
    \State $\mathtt{u} \gets \mathrm{ReLU}(\mathtt{u})$
    \State $\mathtt{u} \gets \mathrm{Dropout}(\mathtt{u})$
    \State $\mathtt{v} \gets W^{(j,2)} \mathtt{u} + b^{(j,2)}$
    \State $\mathtt{v} \gets \mathrm{BN}_{j,2}(\mathtt{v})$
    \State $\mathtt{h^{(j)}} \gets \mathrm{ReLU}\!\big(\mathtt{v} + \mathtt{h^{(j-1)}}\big)$ \Comment{residual skip connection}
  \EndFor
  \State $\mathtt{z} \gets W^{(\mathrm{out})} \mathtt{h^{(BL)}} + b^{(\mathrm{out})}$
  \State \Return $\mathrm{Clamp}\!\big(\mathrm{Sigmoid}(\mathtt{z}),\, \text{min}=\hyperstyle{min\_val},\, \text{max}=1-\hyperstyle{min\_val}\big)$
\EndFunction

\Function{Encode}{$\mathtt{G},\, \mathtt{conv1},\, \mathtt{conv2}$}
  \State $(\mathtt{B}, \mathtt{E_i}, \mathtt{E_a}, \mathtt{batch}) \gets \mathtt{G}$
  \State $\tilde{\mathtt{X}} \gets \texttt{MLP}_{\mathrm{id}}(\mathtt{B})$ \Comment{$\tilde{\mathtt{X}} \in \mathbb{R}^{N \times d_v}$}
  \State $\mathtt{E_{emb}} \gets \texttt{MLP}_{\mathrm{edge}}(\mathtt{E_a})$ \Comment{$\mathtt{E_{emb}} \in \mathbb{R}^{|E| \times d_e}$}
  \State $\mathtt{H} \gets \mathrm{ReLU}\!\big(\mathtt{conv1}(\tilde{\mathtt{X}}, \mathtt{E_i}, \mathtt{E_{emb}})\big)$
  \State $\mathtt{H} \gets \mathrm{ReLU}\!\big(\mathtt{conv2}(\mathtt{H}, \mathtt{E_i}, \mathtt{E_{emb}})\big)$
  \State \Return $\texttt{GlobalMeanPool}(\mathtt{H}, \mathtt{batch})$ \Comment{$\in \mathbb{R}^{b \times d_h}$}
\EndFunction

\Function{ForwardPass}{$\mathtt{BatchDict}$}
  \State $\mathtt{G_s} \gets \mathtt{BatchDict.get}(\texttt{"state\_graph"})$
  \State $\mathtt{s} \gets \textsc{Encode}(\mathtt{G_s},\, \mathtt{s\_conv1},\, \mathtt{s\_conv2})$
  \State $\mathtt{r} \gets \mathtt{s}$ \Comment{no goal, no depth concatenation}
  \State $\mathtt{distance} \gets \textsc{Regressor}(\mathtt{r})$
  \State \Return $\mathtt{distance}$
\EndFunction

\end{algorithmic}
\textbf{Notation:} $d_b$ = bit length of node identifiers, $d_v$ = node embedding dimension, $d_e$ = edge embedding dimension, $d_h$ = hidden dimension in GINE layers, and $b$ = number of graphs in the batch.
\end{algorithm}

\newpage
\section{E-State Representation and Search Strategy}\label{app:experiments:ablation}
\FloatBarrier
In what follows, we present the results for all the experiments of the e-state representation and search strategy study.

We will use the following abbreviations:

\begin{itemize}
    \item \bitmaskRes: e-state representation that uses symbolic ID-based embedding.
    \item \hashedRes: e-state representation that uses hashing-based embedding.
    \item \mappedRes: e-state representation that uses bitmask-based embedding.
    \item \textbf{HFS}: \textbf{Best-First Search}, which we abbreviate as \textbf{HFS} (for Heuristics-First Search) to distinguish it from \textbf{BFS}
    \item \textbf{HFS}$^{*}$.: variant of \textbf{HFS} that augments the heuristics value $h(s)$ predicted by the GNN with the search depth $d(s)$ to form the total score: $f(s) = h(s) + d(s)$
    \item \nodesExp: the number of nodes expanded during search, reflecting the informativeness of the heuristics.
    \item \planLength: the length of the plan found.
    \item \solvingTime: the solving time (in milliseconds).
    \item \IQM: Interquartile Mean, used as an aggregate performance metric.
    \item \IQR: Interquartile Range, reported as a measure of variability.
    \item \myAvg: arithmetic mean, reported as a baseline aggregate metric.
    \item \myStd: standard deviation, used to quantify variability in the data.
    \item \allInstances: indicates that the aggregate value is computed over all instances solved by that approach.
    \item \onlyInCommon: indicates that the aggregate value is computed only over instances solved by all approaches in the comparison.
    \item \myTO: indicates a timeout (after 600 seconds).
    \item \unsolvedColumn: indicates a missing value due to the problem not being solved within the timeout.
\end{itemize}

\subsection{e-State Representations Comparison}
\FloatBarrier
{\small
\begin{longtable}[!ht]{l|ccc|ccc|ccc}
		\centering
		\multirow2{*}{\textbf{Instance Name}} & \multicolumn{3}{c|}{\bitmaskRes} & \multicolumn{3}{c|}{\hashedRes} & \multicolumn{3}{c}{\mappedRes} \\
		\cline{2-10}
		& \planLength & \nodesExp & \solvingTime [ms] & \planLength & \nodesExp & \solvingTime [ms] & \planLength & \nodesExp & \solvingTime [ms] \\
		\hline
		Assemble\_B3-pl\_5 & 5 & 14 & 164 & 5 & 14 & 154 & 5 & 14 & 308 \\
		Assemble\_B5-pl\_5 & 5 & 14 & 223 & 5 & 14 & 371 & 5 & 14 & 475 \\
		Assemble\_B7-pl\_5 & 5 & 14 & 4368 & 5 & 14 & 4790 & 5 & 14 & 9894 \\
		CC\_2\_2\_3-pl\_4 & 4 & 13 & 183 & 5 & 9 & 141 & 4 & 14 & 171 \\
		CC\_2\_2\_3-pl\_6 & 6 & 62 & 424 & 6 & 10 & 106 & 6 & 178 & 1532 \\
		CC\_2\_2\_4-pl\_5 & 7 & 43 & 938 & 9 & 77 & 2507 & 5 & 94 & 3483 \\
\end{longtable}

	\begin{longtable}[!ht]{l|ccc|ccc|ccc}
		\centering
		\small
		\multirow2{*}{\textbf{Instance Name}} & \multicolumn{3}{c|}{\bitmaskRes} & \multicolumn{3}{c|}{\hashedRes} & \multicolumn{3}{c}{\mappedRes} \\
		\cline{2-10}
		& \planLength & \nodesExp & \solvingTime [ms] & \planLength & \nodesExp & \solvingTime [ms] & \planLength & \nodesExp & \solvingTime [ms] \\
		\hline
		Assemble\_B10-pl\_5 & \unsolvedColumn & \unsolvedColumn & \myTO & \unsolvedColumn & \unsolvedColumn & \myTO & \unsolvedColumn & \unsolvedColumn & \myTO\\
		Assemble\_B2-pl\_5 & 5 & 14 & 264 & 5 & 14 & 264 & 5 & 14 & 285 \\
		Assemble\_B4-pl\_5 & 5 & 14 & 275 & 5 & 14 & 332 & 5 & 14 & 330 \\
		Assemble\_B6-pl\_5 & 5 & 14 & 804 & 5 & 14 & 826 & 5 & 14 & 1127 \\
		Assemble\_B8-pl\_5 & 5 & 14 & 24655 & 5 & 14 & 25636 & 5 & 14 & 48340 \\
		Assemble\_B9-pl\_5 & 5 & 14 & 295842 & 5 & 14 & 296625 & \unsolvedColumn & \unsolvedColumn & \myTO\\
\end{longtable}

	\begin{longtable}[!ht]{l|ccc|ccc|ccc}
		\centering
		\small
		\multirow2{*}{\textbf{Instance Name}} & \multicolumn{3}{c|}{\bitmaskRes} & \multicolumn{3}{c|}{\hashedRes} & \multicolumn{3}{c}{\mappedRes} \\
		\cline{2-10}
		& \planLength & \nodesExp & \solvingTime [ms] & \planLength & \nodesExp & \solvingTime [ms] & \planLength & \nodesExp & \solvingTime [ms] \\
		\hline
		Assemble\_B10-pl\_5 & \unsolvedColumn & \unsolvedColumn & \myTO & \unsolvedColumn & \unsolvedColumn & \myTO & \unsolvedColumn & \unsolvedColumn & \myTO\\
		Assemble\_B2-pl\_5 & 5 & 14 & 264 & 5 & 14 & 264 & 5 & 14 & 285 \\
		Assemble\_B3-pl\_5 & 5 & 14 & 164 & 5 & 14 & 154 & 5 & 14 & 308 \\
		Assemble\_B4-pl\_5 & 5 & 14 & 275 & 5 & 14 & 332 & 5 & 14 & 330 \\
		Assemble\_B5-pl\_5 & 5 & 14 & 223 & 5 & 14 & 371 & 5 & 14 & 475 \\
		Assemble\_B6-pl\_5 & 5 & 14 & 804 & 5 & 14 & 826 & 5 & 14 & 1127 \\
\end{longtable}
	
}

\subsection{Symbolic Mapping Representation Search Strategies Comparison}
\FloatBarrier

\begin{longtable}[!ht]{l|ccc|ccc}
	\centering
	\multirow2{*}{\textbf{Instance Name}} & \multicolumn{3}{c|}{\textbf{HFS}$^{*}$} & \multicolumn{3}{c}{\textbf{HFS}} \\
	\cline{2-7}
	& \planLength & \nodesExp & \solvingTime [ms] & \planLength & \nodesExp & \solvingTime [ms] \\
	\hline
	Assemble\_B3-pl\_5 & 5 & 14 & 308 & 5 & 14 & 302 \\
	Assemble\_B5-pl\_5 & 5 & 14 & 475 & 5 & 14 & 448 \\
	Assemble\_B7-pl\_5 & 5 & 14 & 9894 & 5 & 14 & 10020 \\
	CC\_2\_2\_3-pl\_4 & 4 & 14 & 171 & 4 & 9 & 182 \\
	CC\_2\_2\_3-pl\_6 & 6 & 178 & 1532 & 7 & 25 & 291 \\
	CC\_2\_2\_4-pl\_5 & 5 & 94 & 3483 & 12 & 70 & 2787 \\
\end{longtable}

\begin{longtable}[!ht]{l|ccc|ccc}
	\centering
	\multirow2{*}{\textbf{Instance Name}} & \multicolumn{3}{c|}{\textbf{HFS}$^{*}$} & \multicolumn{3}{c}{\textbf{HFS}} \\
	\cline{2-7}
	& \planLength & \nodesExp & \solvingTime [ms] & \planLength & \nodesExp & \solvingTime [ms] \\
	\hline
	Assemble\_B10-pl\_5 & \unsolvedColumn & \unsolvedColumn & \myTO & \unsolvedColumn & \unsolvedColumn & \myTO \\
	Assemble\_B2-pl\_5 & 5 & 14 & 285 & 5 & 14 & 287 \\
	Assemble\_B4-pl\_5 & 5 & 14 & 330 & 5 & 14 & 357 \\
	Assemble\_B6-pl\_5 & 5 & 14 & 1127 & 5 & 14 & 1190 \\
	Assemble\_B8-pl\_5 & 5 & 14 & 48340 & 5 & 14 & 49503 \\
	Assemble\_B9-pl\_5 & \unsolvedColumn & \unsolvedColumn & \myTO & \unsolvedColumn & \unsolvedColumn & \myTO \\
\end{longtable}

\subsection{Hashed-Based Representation Search Strategies Comparison}
\FloatBarrier

\begin{longtable}[!ht]{l|ccc|ccc}
	\centering
	\multirow2{*}{\textbf{Instance Name}} & \multicolumn{3}{c|}{\textbf{HFS}$^{*}$} & \multicolumn{3}{c}{\textbf{HFS}} \\
	\cline{2-7}
	& \planLength & \nodesExp & \solvingTime [ms] & \planLength & \nodesExp & \solvingTime [ms] \\
	\hline
	Assemble\_B3-pl\_5 & 5 & 14 & 154 & 5 & 14 & 249 \\
	Assemble\_B5-pl\_5 & 5 & 14 & 371 & 5 & 14 & 245 \\
	Assemble\_B7-pl\_5 & 5 & 14 & 4790 & 5 & 14 & 4660 \\
	CC\_2\_2\_3-pl\_4 & 5 & 9 & 141 & 7 & 12 & 123 \\
	CC\_2\_2\_3-pl\_6 & 6 & 10 & 106 & 9 & 18 & 246 \\
	CC\_2\_2\_4-pl\_5 & 9 & 77 & 2507 & 13 & 38 & 1221 \\
\end{longtable}

\begin{longtable}[!ht]{l|ccc|ccc}
	\centering
	\multirow2{*}{\textbf{Instance Name}} & \multicolumn{3}{c|}{\textbf{HFS}$^{*}$} & \multicolumn{3}{c}{\textbf{HFS}} \\
	\cline{2-7}
	& \planLength & \nodesExp & \solvingTime [ms] & \planLength & \nodesExp & \solvingTime [ms] \\
	\hline
	Assemble\_B10-pl\_5 & \unsolvedColumn & \unsolvedColumn & \myTO & \unsolvedColumn & \unsolvedColumn & \myTO \\
	Assemble\_B2-pl\_5 & 5 & 14 & 264 & 5 & 14 & 125 \\
	Assemble\_B4-pl\_5 & 5 & 14 & 332 & 5 & 14 & 167 \\
	Assemble\_B6-pl\_5 & 5 & 14 & 826 & 5 & 14 & 689 \\
	Assemble\_B8-pl\_5 & 5 & 14 & 25636 & 5 & 14 & 25949 \\
	Assemble\_B9-pl\_5 & 5 & 14 & 296625 & 5 & 14 & 379798 \\
\end{longtable}

\begin{longtable}[!ht]{l|ccc|ccc}
	\centering
	\multirow2{*}{\textbf{Instance Name}} & \multicolumn{3}{c|}{\textbf{HFS}$^{*}$} & \multicolumn{3}{c}{\textbf{HFS}} \\
	\cline{2-7}
	& \planLength & \nodesExp & \solvingTime [ms] & \planLength & \nodesExp & \solvingTime [ms] \\
	\hline
	Assemble\_B10-pl\_5 & \unsolvedColumn & \unsolvedColumn & \myTO & \unsolvedColumn & \unsolvedColumn & \myTO \\
	Assemble\_B2-pl\_5 & 5 & 14 & 264 & 5 & 14 & 125 \\
	Assemble\_B3-pl\_5 & 5 & 14 & 154 & 5 & 14 & 249 \\
	Assemble\_B4-pl\_5 & 5 & 14 & 332 & 5 & 14 & 167 \\
	Assemble\_B5-pl\_5 & 5 & 14 & 371 & 5 & 14 & 245 \\
	Assemble\_B6-pl\_5 & 5 & 14 & 826 & 5 & 14 & 689 \\
\end{longtable}

\begin{longtable}[!ht]{l|ccc|ccc}
	\centering
	\multirow2{*}{\textbf{Instance Name}} & \multicolumn{3}{c|}{\textbf{HFS}$^{*}$} & \multicolumn{3}{c}{\textbf{HFS}} \\
	\cline{2-7}
	& \planLength & \nodesExp & \solvingTime [ms] & \planLength & \nodesExp & \solvingTime [ms] \\
	\hline
	Assemble\_B10-pl\_5 & \unsolvedColumn & \unsolvedColumn & \myTO & \unsolvedColumn & \unsolvedColumn & \myTO \\
	Assemble\_B2-pl\_5 & 5 & 14 & 285 & 5 & 14 & 287 \\
	Assemble\_B3-pl\_5 & 5 & 14 & 308 & 5 & 14 & 302 \\
	Assemble\_B4-pl\_5 & 5 & 14 & 330 & 5 & 14 & 357 \\
	Assemble\_B5-pl\_5 & 5 & 14 & 475 & 5 & 14 & 448 \\
	Assemble\_B6-pl\_5 & 5 & 14 & 1127 & 5 & 14 & 1190 \\
\end{longtable}

\subsection{Bitmask Representation Search Strategies Comparison}
\FloatBarrier

\begin{longtable}[!ht]{l|ccc|ccc}
	\centering
	\multirow2{*}{\textbf{Instance Name}} & \multicolumn{3}{c|}{\textbf{HFS}$^{*}$} & \multicolumn{3}{c}{\textbf{HFS}} \\
	\cline{2-7}
	& \planLength & \nodesExp & \solvingTime [ms] & \planLength & \nodesExp & \solvingTime [ms] \\
	\hline
	Assemble\_B3-pl\_5 & 5 & 14 & 164 & 5 & 14 & 378 \\
	Assemble\_B5-pl\_5 & 5 & 14 & 223 & 5 & 14 & 323 \\
	Assemble\_B7-pl\_5 & 5 & 14 & 4368 & 5 & 14 & 8500 \\
	CC\_2\_2\_3-pl\_4 & 4 & 13 & 183 & 6 & 12 & 116 \\
	CC\_2\_2\_3-pl\_6 & 6 & 62 & 424 & 7 & 17 & 228 \\
	CC\_2\_2\_4-pl\_5 & 7 & 43 & 938 & 14 & 43 & 847 \\
\end{longtable}

\begin{longtable}[!ht]{l|ccc|ccc}
	\centering
	\multirow2{*}{\textbf{Instance Name}} & \multicolumn{3}{c|}{\textbf{HFS}$^{*}$} & \multicolumn{3}{c}{\textbf{HFS}} \\
	\cline{2-7}
	& \planLength & \nodesExp & \solvingTime [ms] & \planLength & \nodesExp & \solvingTime [ms] \\
	\hline
	Assemble\_B10-pl\_5 & \unsolvedColumn & \unsolvedColumn & \myTO & \unsolvedColumn & \unsolvedColumn & \myTO \\
	Assemble\_B2-pl\_5 & 5 & 14 & 264 & 5 & 14 & 269 \\
	Assemble\_B4-pl\_5 & 5 & 14 & 275 & 5 & 14 & 376 \\
	Assemble\_B6-pl\_5 & 5 & 14 & 804 & 5 & 14 & 1025 \\
	Assemble\_B8-pl\_5 & 5 & 14 & 24655 & 5 & 14 & 35914 \\
	Assemble\_B9-pl\_5 & 5 & 14 & 295842 & 5 & 14 & 465844 \\
\end{longtable}

\begin{longtable}[!ht]{l|ccc|ccc}
	\centering
	\multirow2{*}{\textbf{Instance Name}} & \multicolumn{3}{c|}{\textbf{HFS}$^{*}$} & \multicolumn{3}{c}{\textbf{HFS}} \\
	\cline{2-7}
	& \planLength & \nodesExp & \solvingTime [ms] & \planLength & \nodesExp & \solvingTime [ms] \\
	\hline
	Assemble\_B10-pl\_5 & \unsolvedColumn & \unsolvedColumn & \myTO & \unsolvedColumn & \unsolvedColumn & \myTO \\
	Assemble\_B2-pl\_5 & 5 & 14 & 264 & 5 & 14 & 269 \\
	Assemble\_B3-pl\_5 & 5 & 14 & 164 & 5 & 14 & 378 \\
	Assemble\_B4-pl\_5 & 5 & 14 & 275 & 5 & 14 & 376 \\
	Assemble\_B5-pl\_5 & 5 & 14 & 223 & 5 & 14 & 323 \\
	Assemble\_B6-pl\_5 & 5 & 14 & 804 & 5 & 14 & 1025 \\
\end{longtable}

\newpage

\newpage
\section{Experimental Results}\label{app:experiments}
\FloatBarrier
In what follows, we present the results for all the experiments we conducted.

We will use the following abbreviations:

\begin{itemize}
    \item \GNNres: denotes our primary contribution, \deep equipped with \textbf{A}$\mathbf{^*}$ search and GNN-based heuristics.
    \item \BFSres: \deep using breadth-first search.
    \item \HEFP: a planner that also incorporates heuristics, albeit not derived through learning techniques~\citep{DBLP:conf/prima/FabianoPSP24}. Following \citet{DBLP:conf/prima/FabianoPSP24}, we will use the following to distinguish between the various heuristics used by \HEFP
    \begin{itemize}
        \item \SUB: Heuristics that associates a higher evaluation to e-states that satisfy more sub-goals.
        \item \CPG: Heuristics that emulates the classical Planning Graph by deriving the ``importance'' of each belief formula (its distance from the goal level) and then each e-state is characterized by the sum of the derived belief formulae scores.
    In particular \CPG\ reflects the \emph{hAdd} heuristics in MEP.
    \item \LPG: Heuristics that calculates the score of an e-state by constructing a Planning Graph from it (as initial state) and calculating its length---the shorter the better---of the constructed \epg. This behavior is similar to the one adopted by the heuristics \emph{hFF} in classical planning.
    \item \SPG: This heuristics is simply an execution of \CPG\ on every e-state.
    \end{itemize}
    \item \nodesExp: the number of nodes expanded during search, reflecting the informativeness of the heuristics.
    \item \planLength: the length of the plan found.
    \item \solvingTime: the solving time (in milliseconds).
    \item \IQM: Interquartile Mean, used as an aggregate performance metric.
    \item \IQR: Interquartile Range, reported as a measure of variability.
    \item \myAvg: arithmetic mean, reported as a baseline aggregate metric.
    \item \myStd: standard deviation, used to quantify variability in the data.
    \item \allInstances: indicates that the aggregate value is computed over all instances solved by that approach.
    \item \onlyInCommon: indicates that the aggregate value is computed only over instances solved by all approaches in the comparison.
    \item \myTO: indicates a timeout (after 600 seconds).
    \item \unsolvedColumn: indicates a missing value due to the problem not being solved within the timeout.
\end{itemize}

Since each set of experiments includes both ``Train'' and ``Test'' problems, we report results for each in separate tables.
This distinction is indicated in the table captions using either \textbf{Train} or \textbf{Test}.

\newpage
\subsection{Experimental Setup \#1 -- Standard Benchmarks}
\FloatBarrier

\begin{longtable}[!ht]{l|ccc|ccc}
\centering
\multirow2{*}{\textbf{Instance Name}} & \multicolumn{3}{c|}{\GNNres} & \multicolumn{3}{c}{\BFSres} \\
\cline{2-7}
& \planLength & \nodesExp & \solvingTime [ms] & \planLength & \nodesExp & \solvingTime [ms] \\
\hline
Assemble\_B3-pl\_5 & 5 & 10 & 151 & 5 & 14 & 39 \\
Assemble\_B5-pl\_5 & 5 & 10 & 124 & 5 & 14 & 55 \\
Assemble\_B7-pl\_5 & 5 & 10 & 891 & 5 & 14 & 1057 \\
\hline
\myAvg  $\pm$ \myStd \hfill (\allInstances) & 5 $\pm$ 0 & 10 $\pm$ 0 & 389 $\pm$ 355 & 5 $\pm$ 0 & 14 $\pm$ 0 & 384 $\pm$ 476 \\
\IQM $\pm$ \IQR \hfill (\allInstances) & 5 $\pm$ 0 & 10 $\pm$ 0 & 151 $\pm$ 384 & 5 $\pm$ 0 & 14 $\pm$ 0 & 55 $\pm$ 509 \\
\myAvg  $\pm$ \myStd \hfill (\onlyInCommon) & 5 $\pm$ 0 & 10 $\pm$ 0 & 389 $\pm$ 355 & 5 $\pm$ 0 & 14 $\pm$ 0 & 384 $\pm$ 476 \\
\IQM $\pm$ \IQR \hfill (\onlyInCommon) & 5 $\pm$ 0 & 10 $\pm$ 0 & 151 $\pm$ 384 & 5 $\pm$ 0 & 14 $\pm$ 0 & 55 $\pm$ 509 \\
Solved Instances & \multicolumn{3}{c|}{3/3 (100.00\%)} & \multicolumn{3}{c}{3/3 (100.00\%)}
\\
\caption{Comparison of execution on the Assembly Line domain over the \textbf{Train} instances. The model used by \GNNres has been trained using the instances reported in this Table.}
\label{tab:al_comparison_train}
\end{longtable}

\begin{longtable}[!ht]{l|ccc|ccc}
\centering
\multirow2{*}{\textbf{Instance Name}} & \multicolumn{3}{c|}{\GNNres} & \multicolumn{3}{c}{\BFSres} \\
\cline{2-7}
& \planLength & \nodesExp & \solvingTime [ms] & \planLength & \nodesExp & \solvingTime [ms] \\
\hline
Assemble\_B2-pl\_5 & 5 & 10 & 364 & 5 & 14 & 37 \\
Assemble\_B4-pl\_5 & 5 & 10 & 122 & 5 & 14 & 40 \\
Assemble\_B6-pl\_5 & 5 & 10 & 185 & 5 & 14 & 123 \\
Assemble\_B8-pl\_5 & 5 & 10 & 4270 & 5 & 14 & 5145 \\
Assemble\_B9-pl\_5 & 5 & 10 & 49596 & 5 & 14 & 65910 \\
Assemble\_B10-pl\_5 & \unsolvedColumn & \unsolvedColumn & \myTO & \unsolvedColumn & \unsolvedColumn & \myTO \\
Assemble\_C-pl\_5 & 5 & 10 & 140 & 5 & 14 & 38 \\
\hline
\myAvg  $\pm$ \myStd \hfill (\allInstances) & 5 $\pm$ 0 & 10 $\pm$ 0 & 9113 $\pm$ 18166 & 5 $\pm$ 0 & 14 $\pm$ 0 & 11882 $\pm$ 24233 \\
\IQM $\pm$ \IQR \hfill (\allInstances) & 5 $\pm$ 0 & 10 $\pm$ 0 & 274 $\pm$ 3142 & 5 $\pm$ 0 & 14 $\pm$ 0 & 82 $\pm$ 3851 \\
\myAvg  $\pm$ \myStd \hfill (\onlyInCommon) & 5 $\pm$ 0 & 10 $\pm$ 0 & 9113 $\pm$ 18166 & 5 $\pm$ 0 & 14 $\pm$ 0 & 11882 $\pm$ 24233 \\
\IQM $\pm$ \IQR \hfill (\onlyInCommon) & 5 $\pm$ 0 & 10 $\pm$ 0 & 274 $\pm$ 3142 & 5 $\pm$ 0 & 14 $\pm$ 0 & 82 $\pm$ 3851 \\
Solved Instances & \multicolumn{3}{c|}{6/7 (85.71\%)} & \multicolumn{3}{c}{6/7 (85.71\%)}
\\
\caption{Comparison of execution on the Assembly Line domain over the \textbf{Test} instances. The model used by \GNNres has been trained using the instances reported in the previous \textbf{Train} Table.}
\label{tab:al_comparison_test}
\end{longtable}

\begin{longtable}[!ht]{l|ccc|ccc}
\centering
\multirow2{*}{\textbf{Instance Name}} & \multicolumn{3}{c|}{\GNNres} & \multicolumn{3}{c}{\BFSres} \\
\cline{2-7}
& \planLength & \nodesExp & \solvingTime [ms] & \planLength & \nodesExp & \solvingTime [ms] \\
\hline
CC\_2\_2\_3-pl\_4 & 5 & 8 & 120 & 4 & 17 & 20 \\
CC\_2\_2\_3-pl\_6 & 8 & 67 & 284 & 6 & 287 & 300 \\
CC\_2\_2\_4-pl\_5 & 7 & 22 & 217 & 5 & 170 & 632 \\
CC\_2\_3\_4-pl\_3 & 3 & 19 & 2640 & 3 & 6 & 253 \\
CC\_2\_3\_4-pl\_7 & \unsolvedColumn & \unsolvedColumn & \myTO & 7 & 8907 & 93302 \\
CC\_3\_2\_3-pl\_4 & 5 & 7 & 99 & 4 & 30 & 60 \\
CC\_3\_2\_3-pl\_5 & 8 & 23 & 179 & 5 & 178 & 317 \\
CC\_3\_3\_3-pl\_4 & 7 & 17 & 634 & 4 & 39 & 120 \\
\hline
\myAvg  $\pm$ \myStd \hfill (\allInstances) & 6 $\pm$ 2 & 23 $\pm$ 19 & 596 $\pm$ 851 & 5 $\pm$ 1 & 1204 $\pm$ 2913 & 11876 $\pm$ 30777 \\
\IQM $\pm$ \IQR \hfill (\allInstances) & 6 $\pm$ 2 & 19 $\pm$ 10 & 227 $\pm$ 310 & 4 $\pm$ 1 & 104 $\pm$ 178 & 248 $\pm$ 291 \\
\myAvg  $\pm$ \myStd \hfill (\onlyInCommon) & 6 $\pm$ 2 & 23 $\pm$ 19 & 596 $\pm$ 851 & 4 $\pm$ 1 & 104 $\pm$ 100 & 243 $\pm$ 192 \\
\IQM $\pm$ \IQR \hfill (\onlyInCommon) & 6 $\pm$ 2 & 19 $\pm$ 10 & 227 $\pm$ 310 & 4 $\pm$ 1 & 80 $\pm$ 150 & 224 $\pm$ 218 \\
Solved Instances & \multicolumn{3}{c|}{7/8 (87.50\%)} & \multicolumn{3}{c}{8/8 (100.00\%)}
\\
\caption{Comparison of execution on the Collaboration and Communication domain over the \textbf{Train} instances. The model used by \GNNres has been trained using the instances reported in this Table.}
\label{tab:cc_comparison_train}
\end{longtable}

\begin{longtable}[!ht]{l|ccc|ccc}
\centering
\multirow2{*}{\textbf{Instance Name}} & \multicolumn{3}{c|}{\GNNres} & \multicolumn{3}{c}{\BFSres} \\
\cline{2-7}
& \planLength & \nodesExp & \solvingTime [ms] & \planLength & \nodesExp & \solvingTime [ms] \\
\hline
CC\_2\_2\_3-pl\_3 & 4 & 7 & 155 & 3 & 9 & 13 \\
CC\_2\_2\_3-pl\_5 & 8 & 226 & 1030 & 5 & 78 & 92 \\
CC\_2\_2\_3-pl\_7 & 9 & 271 & 972 & 7 & 1065 & 1111 \\
CC\_2\_2\_3-pl\_8 & 8 & 2373 & 7418 & 8 & 2065 & 2126 \\
CC\_2\_2\_4-pl\_3 & 3 & 11 & 300 & 3 & 6 & 42 \\
CC\_2\_2\_4-pl\_4 & 5 & 10 & 209 & 4 & 24 & 115 \\
\end{longtable}

\begin{longtable}[!ht]{l|ccc|ccc}
\centering
\multirow2{*}{\textbf{Instance Name}} & \multicolumn{3}{c|}{\GNNres} & \multicolumn{3}{c}{\BFSres} \\
\cline{2-7}
& \planLength & \nodesExp & \solvingTime [ms] & \planLength & \nodesExp & \solvingTime [ms] \\
\hline
CoinBox-pl\_3 & 3 & 17 & 139 & 3 & 11 & 22 \\
CoinBox-pl\_6 & 6 & 359 & 1425 & 6 & 912 & 1057 \\
\hline
\myAvg  $\pm$ \myStd \hfill (\allInstances) & 4 $\pm$ 2 & 188 $\pm$ 171 & 782 $\pm$ 643 & 4 $\pm$ 2 & 462 $\pm$ 450 & 540 $\pm$ 518 \\
\IQM $\pm$ \IQR \hfill (\allInstances) & 4 $\pm$ 2 & 188 $\pm$ 171 & 782 $\pm$ 643 & 4 $\pm$ 2 & 462 $\pm$ 450 & 540 $\pm$ 518 \\
\myAvg  $\pm$ \myStd \hfill (\onlyInCommon) & 4 $\pm$ 2 & 188 $\pm$ 171 & 782 $\pm$ 643 & 4 $\pm$ 2 & 462 $\pm$ 450 & 540 $\pm$ 518 \\
\IQM $\pm$ \IQR \hfill (\onlyInCommon) & 4 $\pm$ 2 & 188 $\pm$ 171 & 782 $\pm$ 643 & 4 $\pm$ 2 & 462 $\pm$ 450 & 540 $\pm$ 518 \\
Solved Instances & \multicolumn{3}{c|}{2/2 (100.00\%)} & \multicolumn{3}{c}{2/2 (100.00\%)}
\\
\caption{Comparison of execution on the Coin in the Box domain over the \textbf{Train} instances. The model used by \GNNres has been trained using the instances reported in this Table.}
\label{tab:cb_comparison_train}
\end{longtable}

\begin{longtable}[!ht]{l|ccc|ccc}
\centering
\multirow2{*}{\textbf{Instance Name}} & \multicolumn{3}{c|}{\GNNres} & \multicolumn{3}{c}{\BFSres} \\
\cline{2-7}
& \planLength & \nodesExp & \solvingTime [ms] & \planLength & \nodesExp & \solvingTime [ms] \\
\hline
CoinBox-pl\_2 & 2 & 2 & 84 & 2 & 2 & 8 \\
CoinBox-pl\_5 & 5 & 75 & 279 & 5 & 102 & 118 \\
CoinBox-pl\_7 & 7 & 1721 & 7102 & 7 & 2523 & 3214 \\
\hline
\myAvg  $\pm$ \myStd \hfill (\allInstances) & 5 $\pm$ 2 & 599 $\pm$ 794 & 2488 $\pm$ 3263 & 5 $\pm$ 2 & 876 $\pm$ 1166 & 1113 $\pm$ 1486 \\
\IQM $\pm$ \IQR \hfill (\allInstances) & 5 $\pm$ 2 & 75 $\pm$ 860 & 279 $\pm$ 3509 & 5 $\pm$ 2 & 102 $\pm$ 1260 & 118 $\pm$ 1603 \\
\myAvg  $\pm$ \myStd \hfill (\onlyInCommon) & 5 $\pm$ 2 & 599 $\pm$ 794 & 2488 $\pm$ 3263 & 5 $\pm$ 2 & 876 $\pm$ 1166 & 1113 $\pm$ 1486 \\
\IQM $\pm$ \IQR \hfill (\onlyInCommon) & 5 $\pm$ 2 & 75 $\pm$ 860 & 279 $\pm$ 3509 & 5 $\pm$ 2 & 102 $\pm$ 1260 & 118 $\pm$ 1603 \\
Solved Instances & \multicolumn{3}{c|}{3/3 (100.00\%)} & \multicolumn{3}{c}{3/3 (100.00\%)}
\\
\caption{Comparison of execution on the Coin in the Box domain over the \textbf{Test} instances. The model used by \GNNres has been trained using the instances reported in the previous \textbf{Train} Table.}
\label{tab:cb_comparison_test}
\end{longtable}

\begin{longtable}[!ht]{l|ccc|ccc}
\centering
\multirow2{*}{\textbf{Instance Name}} & \multicolumn{3}{c|}{\GNNres} & \multicolumn{3}{c}{\BFSres} \\
\cline{2-7}
& \planLength & \nodesExp & \solvingTime [ms] & \planLength & \nodesExp & \solvingTime [ms] \\
\hline
Grapevine\_3-pl\_3 & 3 & 18 & 162 & 3 & 31 & 117 \\
Grapevine\_3-pl\_4 & 6 & 38 & 200 & 4 & 160 & 667 \\
Grapevine\_4-pl\_2 & 2 & 3 & 131 & 2 & 7 & 101 \\
Grapevine\_5-pl\_4 & 7 & 68 & 58927 & 4 & 616 & 61732 \\
\hline
\myAvg  $\pm$ \myStd \hfill (\allInstances) & 4 $\pm$ 2 & 32 $\pm$ 24 & 14855 $\pm$ 25445 & 3 $\pm$ 1 & 204 $\pm$ 245 & 15654 $\pm$ 26604 \\
\IQM $\pm$ \IQR \hfill (\allInstances) & 4 $\pm$ 4 & 28 $\pm$ 31 & 181 $\pm$ 14728 & 4 $\pm$ 1 & 96 $\pm$ 249 & 392 $\pm$ 15820 \\
\myAvg  $\pm$ \myStd \hfill (\onlyInCommon) & 4 $\pm$ 2 & 32 $\pm$ 24 & 14855 $\pm$ 25445 & 3 $\pm$ 1 & 204 $\pm$ 245 & 15654 $\pm$ 26604 \\
\IQM $\pm$ \IQR \hfill (\onlyInCommon) & 4 $\pm$ 4 & 28 $\pm$ 31 & 181 $\pm$ 14728 & 4 $\pm$ 1 & 96 $\pm$ 249 & 392 $\pm$ 15820 \\
Solved Instances & \multicolumn{3}{c|}{4/4 (100.00\%)} & \multicolumn{3}{c}{4/4 (100.00\%)}
\\
\caption{Comparison of execution on the Grapevine domain over the \textbf{Train} instances. The model used by \GNNres has been trained using the instances reported in this Table.}
\label{tab:gr_comparison_train}
\end{longtable}

\begin{longtable}[!ht]{l|ccc|ccc}
\centering
\multirow2{*}{\textbf{Instance Name}} & \multicolumn{3}{c|}{\GNNres} & \multicolumn{3}{c}{\BFSres} \\
\cline{2-7}
& \planLength & \nodesExp & \solvingTime [ms] & \planLength & \nodesExp & \solvingTime [ms] \\
\hline
Grapevine\_3-pl\_2 & 2 & 4 & 114 & 2 & 6 & 22 \\
Grapevine\_3-pl\_5 & \unsolvedColumn & \unsolvedColumn & \myTO & 5 & 823 & 2028 \\
Grapevine\_3-pl\_6 & \unsolvedColumn & \unsolvedColumn & \myTO & 6 & 2118 & 4353 \\
Grapevine\_3-pl\_7 & 9 & 10414 & 144193 & 7 & 12008 & 19763 \\
Grapevine\_4-pl\_3 & 6 & 332 & 5342 & 3 & 40 & 612 \\
Grapevine\_4-pl\_4 & 7 & 190 & 10743 & 4 & 233 & 1429 \\
\end{longtable}

\begin{longtable}[!ht]{l|ccc|ccc}
\centering
\multirow2{*}{\textbf{Instance Name}} & \multicolumn{3}{c|}{\GNNres} & \multicolumn{3}{c}{\BFSres} \\
\cline{2-7}
& \planLength & \nodesExp & \solvingTime [ms] & \planLength & \nodesExp & \solvingTime [ms] \\
\hline
SC\_4\_1-pl\_3 & 3 & 3 & 56 & 3 & 4 & 2 \\
SC\_4\_2-pl\_7 & 7 & 729 & 549 & 7 & 136 & 106 \\
SC\_4\_3-pl\_5 & 5 & 31 & 81 & 5 & 11 & 4 \\
SC\_8\_10-pl\_6 & 10 & 24 & 131 & 6 & 113 & 148 \\
SC\_9\_11-pl\_6 & 6 & 19 & 99 & 6 & 17 & 20 \\
SC\_9\_11-pl\_8 & 10 & 15 & 96 & 8 & 65 & 91 \\
SC\_10\_8-pl\_10 & 10 & 10 & 60 & 10 & 10 & 8 \\
SC\_10\_8-pl\_15 & 17 & 23 & 88 & 15 & 96 & 92 \\
SC\_10\_10-pl\_13 & 14 & 33 & 165 & 13 & 52 & 49 \\
\hline
\myAvg  $\pm$ \myStd \hfill (\allInstances) & 9 $\pm$ 4 & 99 $\pm$ 223 & 147 $\pm$ 146 & 8 $\pm$ 4 & 56 $\pm$ 47 & 58 $\pm$ 50 \\
\IQM $\pm$ \IQR \hfill (\allInstances) & 9 $\pm$ 4 & 22 $\pm$ 16 & 99 $\pm$ 50 & 7 $\pm$ 4 & 48 $\pm$ 85 & 52 $\pm$ 84 \\
\myAvg  $\pm$ \myStd \hfill (\onlyInCommon) & 9 $\pm$ 4 & 99 $\pm$ 223 & 147 $\pm$ 146 & 8 $\pm$ 4 & 56 $\pm$ 47 & 58 $\pm$ 50 \\
\IQM $\pm$ \IQR \hfill (\onlyInCommon) & 9 $\pm$ 4 & 22 $\pm$ 16 & 99 $\pm$ 50 & 7 $\pm$ 4 & 48 $\pm$ 85 & 52 $\pm$ 84 \\
Solved Instances & \multicolumn{3}{c|}{9/9 (100.00\%)} & \multicolumn{3}{c}{9/9 (100.00\%)}
\\
\caption{Comparison of execution on the Selective Communication domain over the \textbf{Train} instances. The model used by \GNNres has been trained using the instances reported in this Table.}
\label{tab:sc_comparison_train}
\end{longtable}

\begin{longtable}[!ht]{l|ccc|ccc}
\centering
\multirow2{*}{\textbf{Instance Name}} & \multicolumn{3}{c|}{\GNNres} & \multicolumn{3}{c}{\BFSres} \\
\cline{2-7}
& \planLength & \nodesExp & \solvingTime [ms] & \planLength & \nodesExp & \solvingTime [ms] \\
\hline
SC\_4\_1-pl\_5 & 5 & 26 & 100 & 5 & 11 & 4 \\
SC\_4\_2-pl\_5 & 5 & 549 & 431 & 5 & 22 & 24 \\
SC\_4\_2-pl\_8 & 11 & 1116 & 942 & 8 & 318 & 258 \\
SC\_4\_3-pl\_6 & 6 & 26 & 77 & 6 & 21 & 6 \\
SC\_4\_3-pl\_8 & 8 & 75 & 112 & 8 & 66 & 16 \\
SC\_4\_4-pl\_5 & 5 & 13 & 64 & 5 & 11 & 4 \\
\end{longtable}

\begin{longtable}[!ht]{l|ccc|ccc}
\centering
\multirow2{*}{\textbf{Instance Name}} & \multicolumn{3}{c|}{\GNNres} & \multicolumn{3}{c}{\BFSres} \\
\cline{2-7}
& \planLength & \nodesExp & \solvingTime [ms] & \planLength & \nodesExp & \solvingTime [ms] \\
\hline
SC\_R\_10\_10-pl\_6 & 6 & 298 & 1987 & 6 & 333 & 1635 \\
SC\_R\_10\_10-pl\_7 & 7 & 589 & 3870 & 7 & 792 & 3951 \\
SC\_R\_10\_10-pl\_10 & 10 & 1977 & 11786 & 10 & 2431 & 11831 \\
\hline
\myAvg  $\pm$ \myStd \hfill (\allInstances) & 8 $\pm$ 2 & 955 $\pm$ 733 & 5881 $\pm$ 4246 & 8 $\pm$ 2 & 1185 $\pm$ 901 & 5806 $\pm$ 4364 \\
\IQM $\pm$ \IQR \hfill (\allInstances) & 7 $\pm$ 2 & 589 $\pm$ 840 & 3870 $\pm$ 4900 & 7 $\pm$ 2 & 792 $\pm$ 1049 & 3951 $\pm$ 5098 \\
\myAvg  $\pm$ \myStd \hfill (\onlyInCommon) & 8 $\pm$ 2 & 955 $\pm$ 733 & 5881 $\pm$ 4246 & 8 $\pm$ 2 & 1185 $\pm$ 901 & 5806 $\pm$ 4364 \\
\IQM $\pm$ \IQR \hfill (\onlyInCommon) & 7 $\pm$ 2 & 589 $\pm$ 840 & 3870 $\pm$ 4900 & 7 $\pm$ 2 & 792 $\pm$ 1049 & 3951 $\pm$ 5098 \\
Solved Instances & \multicolumn{3}{c|}{3/3 (100.00\%)} & \multicolumn{3}{c}{3/3 (100.00\%)}
\\
\caption{Comparison of execution on the Selective Communication Enriched domain over the \textbf{Train} instances. The model used by \GNNres has been trained using the instances reported in this Table.}
\label{tab:sr_comparison_train}
\end{longtable}

\begin{longtable}[!ht]{l|ccc|ccc}
\centering
\multirow2{*}{\textbf{Instance Name}} & \multicolumn{3}{c|}{\GNNres} & \multicolumn{3}{c}{\BFSres} \\
\cline{2-7}
& \planLength & \nodesExp & \solvingTime [ms] & \planLength & \nodesExp & \solvingTime [ms] \\
\hline
SC\_R\_10\_10-pl\_2 & 2 & 6 & 125 & 2 & 5 & 45 \\
SC\_R\_10\_10-pl\_3 & 3 & 10 & 109 & 3 & 17 & 88 \\
SC\_R\_10\_10-pl\_9 & 9 & 1248 & 7447 & 9 & 1112 & 6079 \\
SC\_R\_10\_10-pl\_13 & 13 & 17305 & 106761 & 13 & 22625 & 120270 \\
SC\_R\_10\_10-pl\_14 & 14 & 38714 & 239252 & 14 & 47255 & 239197 \\
SC\_R\_10\_10-pl\_17 & \unsolvedColumn & \unsolvedColumn & \myTO & \unsolvedColumn & \unsolvedColumn & \myTO \\
\hline
\myAvg  $\pm$ \myStd \hfill (\allInstances) & 8 $\pm$ 5 & 11457 $\pm$ 15123 & 70739 $\pm$ 93461 & 8 $\pm$ 5 & 14203 $\pm$ 18642 & 73136 $\pm$ 94840 \\
\IQM $\pm$ \IQR \hfill (\allInstances) & 8 $\pm$ 10 & 6188 $\pm$ 17295 & 38111 $\pm$ 106636 & 8 $\pm$ 10 & 7918 $\pm$ 22608 & 42146 $\pm$ 120182 \\
\myAvg  $\pm$ \myStd \hfill (\onlyInCommon) & 8 $\pm$ 5 & 11457 $\pm$ 15123 & 70739 $\pm$ 93461 & 8 $\pm$ 5 & 14203 $\pm$ 18642 & 73136 $\pm$ 94840 \\
\IQM $\pm$ \IQR \hfill (\onlyInCommon) & 8 $\pm$ 10 & 6188 $\pm$ 17295 & 38111 $\pm$ 106636 & 8 $\pm$ 10 & 7918 $\pm$ 22608 & 42146 $\pm$ 120182 \\
Solved Instances & \multicolumn{3}{c|}{5/6 (83.33\%)} & \multicolumn{3}{c}{5/6 (83.33\%)}
\\
\caption{Comparison of execution on the Selective Communication Enriched domain over the \textbf{Test} instances. The model used by \GNNres has been trained using the instances reported in the previous \textbf{Train} Table.}
\label{tab:sr_comparison_test}
\end{longtable}

\begin{longtable}[!ht]{l|ccc|ccc}
\centering
\multirow2{*}{\textbf{Instance Name}} & \multicolumn{3}{c|}{\GNNres} & \multicolumn{3}{c}{\BFSres} \\
\cline{2-7}
& \planLength & \nodesExp & \solvingTime [ms] & \planLength & \nodesExp & \solvingTime [ms] \\
\hline
Assemble\_B3-pl\_5 & 5 & 10 & 151 & 5 & 14 & 39 \\
Assemble\_B5-pl\_5 & 5 & 10 & 124 & 5 & 14 & 55 \\
Assemble\_B7-pl\_5 & 5 & 10 & 891 & 5 & 14 & 1057 \\
CC\_2\_2\_3-pl\_4 & 5 & 8 & 120 & 4 & 17 & 20 \\
CC\_2\_2\_3-pl\_6 & 8 & 67 & 284 & 6 & 287 & 300 \\
CC\_2\_2\_4-pl\_5 & 7 & 22 & 217 & 5 & 170 & 632 \\
\end{longtable}

\begin{longtable}[!ht]{l|ccc|ccc}
\centering
\multirow2{*}{\textbf{Instance Name}} & \multicolumn{3}{c|}{\GNNres} & \multicolumn{3}{c}{\BFSres} \\
\cline{2-7}
& \planLength & \nodesExp & \solvingTime [ms] & \planLength & \nodesExp & \solvingTime [ms] \\
\hline
Assemble\_B10-pl\_5 & \unsolvedColumn & \unsolvedColumn & \myTO & \unsolvedColumn & \unsolvedColumn & \myTO \\
Assemble\_B2-pl\_5 & 5 & 10 & 364 & 5 & 14 & 37 \\
Assemble\_B4-pl\_5 & 5 & 10 & 122 & 5 & 14 & 40 \\
Assemble\_B6-pl\_5 & 5 & 10 & 185 & 5 & 14 & 123 \\
Assemble\_B8-pl\_5 & 5 & 10 & 4270 & 5 & 14 & 5145 \\
Assemble\_B9-pl\_5 & 5 & 10 & 49596 & 5 & 14 & 65910 \\
\end{longtable}

\newpage
\subsection{Experimental Setup \#2 -- Knowledge Transfer}
\FloatBarrier

Experimental Setup \#2 serves as the Knowledge Transfer experimental results.

{}

To accomplish this, we trained the model \modelMinGR on the \textbf{Collaboration and Communication} and \textbf{Grapevine} domains, and tested it against all other domains.
To evaluate this model against the other heuristics, we also included the \textbf{Epistemic Gossip} domain to expand the testing set for the various heuristics.

\begin{longtable}[!ht]{l|ccc|ccc}
\centering
\multirow2{*}{\textbf{Instance Name}} & \multicolumn{3}{c|}{\GNNres} & \multicolumn{3}{c}{\BFSres} \\
\cline{2-7}
& \planLength & \nodesExp & \solvingTime [ms] & \planLength & \nodesExp & \solvingTime [ms] \\
\hline
CC\_2\_2\_3-pl\_4 & 4 & 14 & 129 & 4 & 17 & 35 \\
CC\_2\_2\_3-pl\_6 & 6 & 178 & 997 & 6 & 287 & 705 \\
CC\_2\_2\_4-pl\_5 & 5 & 94 & 1915 & 5 & 170 & 1262 \\
CC\_2\_3\_4-pl\_3 & 3 & 19 & 2510 & 3 & 6 & 446 \\
CC\_2\_3\_4-pl\_7 & 7 & 4153 & 341064 & 7 & 8907 & 171616 \\
CC\_3\_2\_3-pl\_4 & 4 & 47 & 418 & 4 & 30 & 91 \\
CC\_3\_2\_3-pl\_5 & 5 & 136 & 1331 & 5 & 178 & 549 \\
CC\_3\_3\_3-pl\_4 & 4 & 39 & 690 & 4 & 39 & 197 \\
Grapevine\_3-pl\_3 & 3 & 21 & 266 & 3 & 31 & 177 \\
Grapevine\_3-pl\_4 & 4 & 113 & 1863 & 4 & 160 & 1374 \\
Grapevine\_4-pl\_2 & 2 & 4 & 398 & 2 & 7 & 160 \\
Grapevine\_5-pl\_4 & 4 & 430 & 191371 & 4 & 616 & 120092 \\
\hline
\myAvg  $\pm$ \myStd \hfill (\allInstances) & 4 $\pm$ 1 & 437 $\pm$ 1126 & 45246 $\pm$ 103440 & 4 $\pm$ 1 & 871 $\pm$ 2429 & 24725 $\pm$ 55183 \\
\IQM $\pm$ \IQR \hfill (\allInstances) & 4 $\pm$ 1 & 75 $\pm$ 126 & 1202 $\pm$ 1651 & 4 $\pm$ 1 & 101 $\pm$ 178 & 556 $\pm$ 1117 \\
\myAvg  $\pm$ \myStd \hfill (\onlyInCommon) & 4 $\pm$ 1 & 437 $\pm$ 1126 & 45246 $\pm$ 103440 & 4 $\pm$ 1 & 871 $\pm$ 2429 & 24725 $\pm$ 55183 \\
\IQM $\pm$ \IQR \hfill (\onlyInCommon) & 4 $\pm$ 1 & 75 $\pm$ 126 & 1202 $\pm$ 1651 & 4 $\pm$ 1 & 101 $\pm$ 178 & 556 $\pm$ 1117 \\
Solved Instances & \multicolumn{3}{c|}{12/12 (100.00\%)} & \multicolumn{3}{c}{12/12 (100.00\%)}
\\
\caption{Comparison of execution on the \textbf{Train} instances of the domains used for the model \modelMinGR.}
\label{tab:batch4_CC-Grapevine_comparison_train}
\end{longtable}

{%
\begin{longtable}[!ht]{l|ccc|ccc}
\centering
\multirow2{*}{\textbf{Instance Name}} & \multicolumn{3}{c|}{\GNNres} & \multicolumn{3}{c}{\BFSres} \\
\cline{2-7}
& \planLength & \nodesExp & \solvingTime [ms] & \planLength & \nodesExp & \solvingTime [ms] \\
\hline
CC\_2\_2\_3-pl\_3 & 3 & 9 & 123 & 3 & 9 & 12 \\
CC\_2\_2\_3-pl\_5 & 5 & 77 & 424 & 5 & 78 & 117 \\
CC\_2\_2\_3-pl\_7 & 7 & 558 & 2502 & 7 & 1065 & 1789 \\
CC\_2\_2\_3-pl\_8 & 8 & 2539 & 15968 & 8 & 2065 & 3288 \\
CC\_2\_2\_4-pl\_3 & 3 & 11 & 184 & 3 & 6 & 39 \\
CC\_2\_2\_4-pl\_4 & 4 & 23 & 332 & 4 & 24 & 155 \\
\end{longtable}}

\newpage

\clearpage
\subsection*{\HEFP's Individual Heuristics against Model \modelMinGR}
\FloatBarrier

{\scriptsize
\begin{longtable}[!ht]{l|ccc|ccc|cccc}
\centering
\multirow{2}{*}{\textbf{Instance Name}} & \multicolumn{3}{c|}{\GNNres} & \multicolumn{3}{c|}{\BFSres} & \multicolumn{4}{c}{\HEFP \CPG\ only} \\
\cline{2-11}
& \planLength & \nodesExp & \solvingTime [ms] & \planLength & \nodesExp & \solvingTime [ms] & \planLength & \nodesExp & \solvingTime [ms] & Heur. \\
\hline
Assemble\_B2-pl\_5 & 5 & 14 & 140 & 5 & 14 & 47 & 5 & 5 & 97 & \CPG \\
Assemble\_B4-pl\_5 & 5 & 14 & 243 & 5 & 14 & 77 & 5 & 5 & 114 & \CPG \\
Assemble\_B6-pl\_5 & 5 & 14 & 760 & 5 & 14 & 461 & 5 & 5 & 1852 & \CPG \\
Assemble\_B8-pl\_5 & 5 & 14 & 31750 & 5 & 14 & 24112 & 5 & 5 & 147496 & \CPG \\
Assemble\_B9-pl\_5 & 5 & 14 & 585443 & 5 & 14 & 390542 & \unsolvedColumn & \unsolvedColumn & \myTO & \unsolvedColumn \\
Assemble\_B10-pl\_5 & \unsolvedColumn & \unsolvedColumn & \myTO & \unsolvedColumn & \unsolvedColumn & \myTO & \unsolvedColumn & \unsolvedColumn & \myTO & \unsolvedColumn \\
\end{longtable}}

\clearpage
{\scriptsize
\begin{longtable}[!ht]{l|ccc|ccc|cccc}
\centering
\multirow{2}{*}{\textbf{Instance Name}} & \multicolumn{3}{c|}{\GNNres} & \multicolumn{3}{c|}{\BFSres} & \multicolumn{4}{c}{\HEFP \LPG\ only} \\
\cline{2-11}
& \planLength & \nodesExp & \solvingTime [ms] & \planLength & \nodesExp & \solvingTime [ms] & \planLength & \nodesExp & \solvingTime [ms] & Heur. \\
\hline
Assemble\_B2-pl\_5 & 5 & 14 & 140 & 5 & 14 & 47 & 5 & 8 & 164 & \LPG \\
Assemble\_B4-pl\_5 & 5 & 14 & 243 & 5 & 14 & 77 & 5 & 8 & 218 & \LPG \\
Assemble\_B6-pl\_5 & 5 & 14 & 760 & 5 & 14 & 461 & 5 & 8 & 2995 & \LPG \\
Assemble\_B8-pl\_5 & 5 & 14 & 31750 & 5 & 14 & 24112 & 5 & 8 & 241692 & \LPG \\
Assemble\_B9-pl\_5 & 5 & 14 & 585443 & 5 & 14 & 390542 & \unsolvedColumn & \unsolvedColumn & \myTO & \unsolvedColumn \\
Assemble\_B10-pl\_5 & \unsolvedColumn & \unsolvedColumn & \myTO & \unsolvedColumn & \unsolvedColumn & \myTO & \unsolvedColumn & \unsolvedColumn & \myTO & \unsolvedColumn \\
\end{longtable}}

\clearpage
{\scriptsize
\begin{longtable}[!ht]{l|ccc|ccc|cccc}
\centering
\multirow{2}{*}{\textbf{Instance Name}} & \multicolumn{3}{c|}{\GNNres} & \multicolumn{3}{c|}{\BFSres} & \multicolumn{4}{c}{\HEFP \SPG\ only} \\
\cline{2-11}
& \planLength & \nodesExp & \solvingTime [ms] & \planLength & \nodesExp & \solvingTime [ms] & \planLength & \nodesExp & \solvingTime [ms] & Heur. \\
\hline
Assemble\_B2-pl\_5 & 5 & 14 & 140 & 5 & 14 & 47 & 5 & 8 & 176 & \SPG \\
Assemble\_B4-pl\_5 & 5 & 14 & 243 & 5 & 14 & 77 & 5 & 8 & 209 & \SPG \\
Assemble\_B6-pl\_5 & 5 & 14 & 760 & 5 & 14 & 461 & 5 & 8 & 3140 & \SPG \\
Assemble\_B8-pl\_5 & 5 & 14 & 31750 & 5 & 14 & 24112 & 5 & 8 & 239144 & \SPG \\
Assemble\_B9-pl\_5 & 5 & 14 & 585443 & 5 & 14 & 390542 & \unsolvedColumn & \unsolvedColumn & \myTO & \unsolvedColumn \\
Assemble\_B10-pl\_5 & \unsolvedColumn & \unsolvedColumn & \myTO & \unsolvedColumn & \unsolvedColumn & \myTO & \unsolvedColumn & \unsolvedColumn & \myTO & \unsolvedColumn \\
\end{longtable}}

\clearpage

{\scriptsize
\begin{longtable}[!ht]{l|ccc|ccc|cccc}
\centering
\multirow{2}{*}{\textbf{Instance Name}} & \multicolumn{3}{c|}{\GNNres} & \multicolumn{3}{c|}{\BFSres} & \multicolumn{4}{c}{\HEFP \SUB\ only} \\
\cline{2-11}
& \planLength & \nodesExp & \solvingTime [ms] & \planLength & \nodesExp & \solvingTime [ms] & \planLength & \nodesExp & \solvingTime [ms] & Heur. \\
\hline
Assemble\_B2-pl\_5 & 5 & 14 & 140 & 5 & 14 & 47 & 5 & 10 & 274 & \SUB \\
Assemble\_B4-pl\_5 & 5 & 14 & 243 & 5 & 14 & 77 & 5 & 10 & 287 & \SUB \\
Assemble\_B6-pl\_5 & 5 & 14 & 760 & 5 & 14 & 461 & 5 & 10 & 921 & \SUB \\
Assemble\_B8-pl\_5 & 5 & 14 & 31750 & 5 & 14 & 24112 & 5 & 10 & 40687 & \SUB \\
Assemble\_B9-pl\_5 & 5 & 14 & 585443 & 5 & 14 & 390542 & 5 & 10 & 594127 & \SUB \\
Assemble\_B10-pl\_5 & \unsolvedColumn & \unsolvedColumn & \myTO & \unsolvedColumn & \unsolvedColumn & \myTO & \unsolvedColumn & \unsolvedColumn & \myTO & \unsolvedColumn \\
\end{longtable}}

\FloatBarrier
\newpage
\section{Overhead of Single-Query GNN-Based Estimation}\label{app:batch-estimate}

Although our current implementation evaluates the GNN heuristic one state at a time, we can estimate the potential impact of batching using aggregate runtime statistics.

From Table~\ref{tab:aggregate}, \BFSres requires $1384/389 \approx 3.56$ ms (time per expanded node), whereas the GNN-based configuration requires $4116/288 \approx 14.29$ ms per node. The difference,
$\beta \approx 10.73$ ms, represents the effective overhead of a single-query GNN evaluation.

Assuming that batching affects only this scoring component, we model the per-node cost under batch size $B$ as
\[
r(B) = r_{\text{BFS}} + \frac{\beta}{B},
\]
where $r_{\text{BFS}} = 3.56$ ms denotes the planner-dominated per-node cost. For $B=1$, this yields
$r(1) = r_{\text{BFS}} + \beta \approx 14.29$ ms, matching the observed single-query runtime.

The total projected runtime under batching is then
\[
T(B) = N_{\text{GNN}} \cdot r(B),
\]
where $N_{\text{GNN}}$ is the number of nodes expanded by \GNNres.

Figure~\ref{fig:batch_scalability_all} reports the resulting estimates for
$B \in \{1, 8, 16, 32, 64, 128, 256, 512\}$. The results show that batching rapidly amortizes the GNN overhead: with $B=32$, the projected runtime decreases from $4116$ ms to approximately $1123$ ms, approaching the planner-dominated regime. This supports our choice of expanded nodes as the primary evaluation metric, as heuristic informativeness remains invariant under such engineering optimizations.

Finally, we quantify the relative speed-up induced by batching. Given the projected runtime $T(B)$, we define the speed-up with respect to single-query inference ($B=1$) as
\[
S_{\text{rel}}(B) := \frac{T(1)}{T(B)}
= \frac{r(1)}{r(B)}
= \frac{r_{\text{BFS}} + \beta}{r_{\text{BFS}} + \beta/B}.
\]
Note that $S_{\text{rel}}(B) \le B$, with diminishing returns as $r(B)$ approaches the planner-dominated baseline $r_{\text{BFS}}$.

\begin{figure*}[!ht]
    \centering
    
    \begin{subfigure}{0.32\textwidth}
        \centering
        \begin{tikzpicture}
        \begin{axis}[
            width=\linewidth,
            height=0.85\linewidth,
            xmode=log,
            log basis x=2,
            xlabel={Batch size $B$},
            ylabel={Per-node overhead [ms]},
            xtick={1,8,16,32,64,128,256,512},
            xticklabels={1,8,16,32,64,128,256,512},
            grid=both,
            ymin=3, ymax=15,
            legend style={at={(0.98,0.98)},anchor=north east, font=\scriptsize},
            tick label style={font=\scriptsize},
            label style={font=\scriptsize},
        ]
        \addplot[thick, mark=*] coordinates {
            (1,14.29) (8,4.90) (16,4.23) (32,3.90)
            (64,3.73) (128,3.64) (256,3.60) (512,3.58)
        };
        \addplot[dashed, thick] coordinates {(1,3.56) (512,3.56)};
        \legend{$r(B)$, $r_{\text{BFS}}$}
        \end{axis}
        \end{tikzpicture}
        \caption{Per-node cost.}
        \label{fig:batch_per_node}
    \end{subfigure}
    \hfill
    \begin{subfigure}{0.32\textwidth}
        \centering
        \begin{tikzpicture}
        \begin{axis}[
            width=\linewidth,
            height=0.85\linewidth,
            xmode=log,
            log basis x=2,
            xlabel={Batch size $B$},
            ylabel={Projected time [ms]},
            xtick={1,8,16,32,64,128,256,512},
            xticklabels={1,8,16,32,64,128,256,512},
            grid=both,
            ymin=900, ymax=4300,
            legend style={at={(0.98,0.98)},anchor=north east, font=\scriptsize},
            tick label style={font=\scriptsize},
            label style={font=\scriptsize},
        ]
        \addplot[thick, mark=*] coordinates {
            (1,4116) (8,1411) (16,1218) (32,1123)
            (64,1074) (128,1050) (256,1037) (512,1031)
        };
        \addplot[dashed, thick] coordinates {(1,1025) (512,1025)};
        \legend{$N_{\text{GNN}}\,r(B)$, $N_{\text{GNN}}\,r_{\text{BFS}}$}
        \end{axis}
        \end{tikzpicture}
        \caption{Projected runtime.}
        \label{fig:batch_runtime}
    \end{subfigure}
    \hfill
    \begin{subfigure}{0.32\textwidth}
        \centering
        \begin{tikzpicture}
        \begin{axis}[
            width=\linewidth,
            height=0.85\linewidth,
            xmode=log,
            log basis x=2,
            xlabel={Batch size $B$},
            ylabel={Speed-up $S_{\text{rel}}(B)$},
            xtick={1,8,16,32,64,128,256,512},
            xticklabels={1,8,16,32,64,128,256,512},
            grid=both,
            ymin=1, ymax=5,
            legend style={at={(0.02,0.98)},anchor=north west, font=\scriptsize},
            tick label style={font=\scriptsize},
            label style={font=\scriptsize},
        ]
        \addplot[thick, mark=*] coordinates {
            (1,1.00) (8,2.92) (16,3.38) (32,3.67)
            (64,3.83) (128,3.92) (256,3.97) (512,3.99)
        };
        \legend{$S_{\text{rel}}(B)=T(1)/T(B)$}
        \end{axis}
        \end{tikzpicture}
        \caption{Relative speed-up.}
        \label{fig:batch_speedup}
    \end{subfigure}
    
    \caption{
   Projected batching scalability for the GNN-based heuristic evaluator.
Per-node cost is modeled as $r(B)=r_{\text{BFS}}+\beta/B$, and total runtime as
$T(B)=N_{\text{GNN}}\cdot r(B)$.
The resulting relative speed-up is
\(
S_{\text{rel}}(B)=\frac{T(1)}{T(B)}
=\frac{r_{\text{BFS}}+\beta}{r_{\text{BFS}}+\beta/B}.
\)
Batching amortizes single-query GNN overhead, with diminishing returns as costs
approach the planner-dominated regime.
}
    \label{fig:batch_scalability_all}
\end{figure*}

}

    \fi

\end{document}